\documentclass{article}

\usepackage{microtype}
\usepackage{graphicx}
\usepackage{booktabs} % for professional tables

\usepackage{hyperref}

\usepackage{xspace}
\usepackage{multirow}
\usepackage{caption}
\usepackage{subcaption}
\usepackage{amsmath}
\usepackage{amsfonts}
\usepackage{amssymb}
\usepackage{enumitem}

\newcommand{\logicept}{{\sc LogicE}\xspace}
\newcommand{\logice}{{\sc LogicE}\xspace}
\newcommand{\logiceb}{{~$+$bounds}\xspace}
\newcommand{\betae}{{\sc BetaE}\xspace}
\newcommand{\qb}{{\sc q2b}\xspace}
\newcommand{\gqe}{{\sc gqe}\xspace}

\makeatletter
\newcommand\footnoteref[1]{\protected@xdef\@thefnmark{\ref{#1}}\@footnotemark}
\makeatother

\usepackage[accepted]{icml2021}

\icmltitlerunning{Logic Embeddings for Complex Query Answering}

\begin{document}

\twocolumn[
\icmltitle{Logic Embeddings for Complex Query Answering}



\icmlsetsymbol{equal}{*}

\begin{icmlauthorlist}
\icmlauthor{Francois Luus}{ibmres}
\icmlauthor{Prithviraj Sen}{ibmres}
\icmlauthor{Pavan Kapanipathi}{ibmres}
\icmlauthor{Ryan Riegel}{ibmres}
\icmlauthor{Ndivhuwo Makondo}{ibmres}
\icmlauthor{Thabang Lebese}{ibmres}
\icmlauthor{Alexander Gray}{ibmres}
\end{icmlauthorlist}

\icmlaffiliation{ibmres}{IBM Research}

\icmlcorrespondingauthor{}{https://github.com/francoisluus/KGReasoning}

\icmlkeywords{Query answering, knowledge graph, embedding, link prediction}

\vskip 0.3in
]



\printAffiliationsAndNotice{}  % leave blank if no need to mention equal contribution

\begin{abstract}
Answering logical queries over incomplete knowledge bases is challenging because: 1) it calls for implicit link prediction, and 2) brute force answering of existential first-order logic queries is exponential in the number of existential variables. Recent work of query embeddings provides fast querying, but most approaches model set logic with closed regions, so lack negation. Query embeddings that do support negation use densities that suffer drawbacks: 1) only improvise logic, 2) use expensive distributions, and 3) poorly model answer uncertainty. In this paper, we propose Logic Embeddings, a new approach to embedding complex queries that uses Skolemisation to eliminate existential variables for efficient querying. It supports negation, but improves on density approaches: 1) integrates well-studied t-norm logic and directly evaluates satisfiability, 2) simplifies modeling with truth values, and 3) models uncertainty with truth bounds. Logic Embeddings are competitively fast and accurate in query answering over large, incomplete knowledge graphs, outperform on negation queries, and in particular, provide improved modeling of answer uncertainty as evidenced by a superior correlation between answer set size and embedding entropy.
\end{abstract}

\section{Introduction}
\begin{figure*}[t!]
% \vskip 0.2in
\begin{center}
\centerline{\includegraphics[width=\textwidth,trim={0cm 8.2cm 0.2cm 0cm},clip]{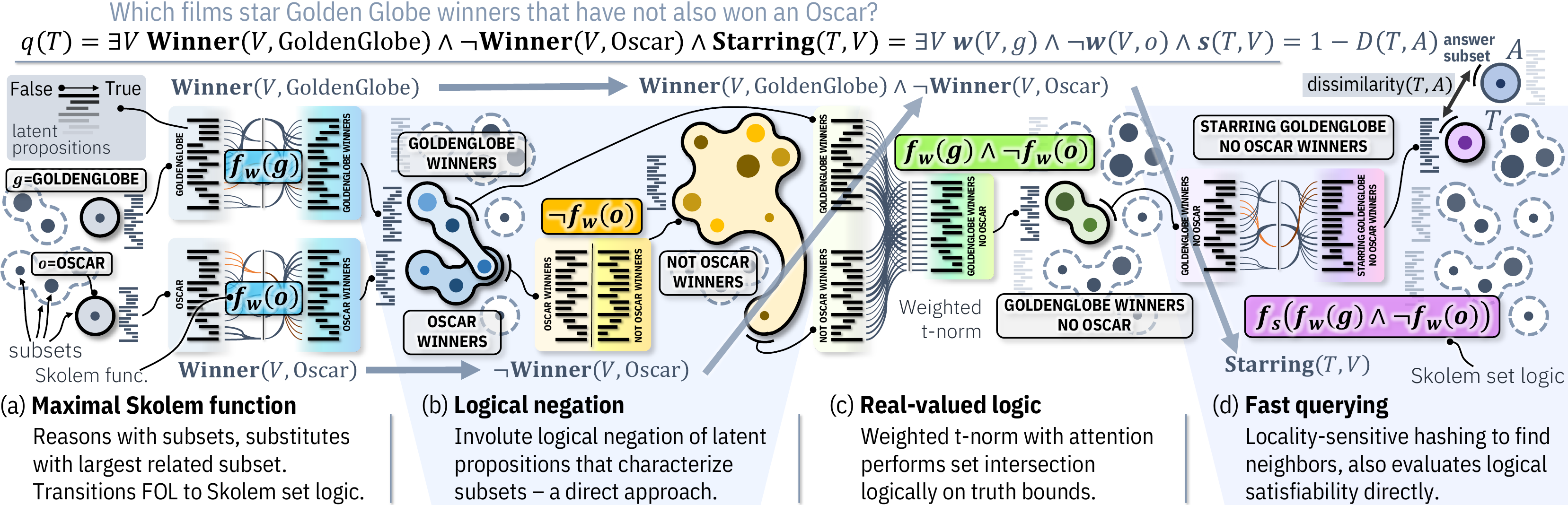}}
\vspace*{-2mm}
\caption{Logic embeddings perform real-valued logic on latent propositions (latents), an array of truth bounds that describes any subset of entities.
% , and incorporate knowledge with Skolem functions to answer \textit{``Which films star Golden Globe winners that have not also won an Oscar?"}:
(a) A learnt Skolem function maps latents of singleton \textit{Oscar} to latents of maximal subset of \textit{Oscar winners}, and similarly for \textit{GoldenGlobe};
(b) Complement of a subset is logical negation of latents that identify non-\textit{Oscar winners};
(c) Intersection of subsets is logical conjunction of latents that identify \textit{``Golden Globe winners that have not also won an Oscar"};
(d) $q(T)=1-D(T,A)$ measures logic satisfiability of candidate answer set directly, while nearest neighbors to intermediate embeddings can provide some explainability.
}
\label{example1}
\end{center}
\vskip -0.3in
\end{figure*}

Reasoning over knowledge bases is fundamental to Artificial Intelligence, but still challenging since most knowledge graphs (KGs) 
such as DBpedia \cite{bizer2009dbpedia}, Freebase \cite{bollacker2008freebase}, and NELL \cite{carlson2010toward} 
are often large and incomplete.
Answering complex queries is an important use of KGs, but missing facts makes queries unanswerable under normal inference.
Figure~\ref{example1} shows an example of handling a logic query representing the natural language question \textit{``Which films star Golden Globe winners that have not also won an Oscar?"}. 
Answering this query involves multiple steps of KG traversal and existential first-order logic (FOL) operations, each producing intermediate entities.
We consider queries involving missing facts, which means there is uncertainty about these intermediates that complicates the task.
Two main approaches to answering such multi-hop queries involving missing facts are (i) sequential path search and (ii) query embeddings.
Sequential path search grows exponentially in the number of hops, and requires approaches like reinforcement learning \cite{das2017go} or beam search \cite{arakelyan2020complex} that have to explicitly track intermediate entities.
% KG embeddings are popular for predicting facts, learning entities as vectors and relations between them as functions in vector space, like translation \cite{bordes2013translating} or rotation \cite{sun2019rotate}.
% Link prediction uncovers similar behavior of entities, and semantic similarity between relations (e.g. birthplace predicts nationality).
% Path queries involve multi-hop reasoning (e.g. country of birthcity of person), where compositional learning embeds queries close to answer entities with fast (sublinear) neighbor search \cite{guu2015traversing}.
% In contrast, sequential path search grows exponentially in the number of hops, and requires approaches like reinforcement learning \cite{das2017go} or beam search \cite{arakelyan2020complex} that have to explicitly track intermediate entities.
Query embeddings prefer composition over search, for fast (sublinear) inference and tractable scaling to more complex queries.
While relation functions have to learn knowledge, composition can otherwise use inductive bias to model logic operators directly to alleviate learning difficulty.

Query embeddings need to (a) predict missing knowledge, (b) model logic operations, and (c) model answer uncertainty.
Query2Box models conjunction as intersection of boxes,
but is unable to model negation as the complement of a closed region is not closed \cite{ren2020query2box}.
Beta embeddings of \cite{ren2020beta} model conjunction as weighted interpolation of Beta distributions and negation as inversion of density, but improvise logic and depend on neural versions of logic conjunction for better accuracy.

% KG embeddings typically 
\cite{hamilton2018embedding} models entities as points so are unable to naturally express uncertainty, while Query2Box uses poorly differentiable geometric shapes unsuited to uncertainty calculations.
% KG2E \cite{he2015learning} instead use Gaussian densities to model inclusion and entailment of concept pairs, but do not extend to complex multi-hop queries.
Beta embeddings naturally model uncertainty with densities
and do support complex query embedding, although its densities have no closed form and entropy calculations are expensive.

Beta embeddings \cite{ren2020beta} are the first query embedding that supports negation and models uncertainty, however it
(1) abruptly converts first-order logic to set logic,
(2) only improvises set logic with densities,
(3) requires expensive Beta distribution (no closed form), and
(4) dissimilarity uses divergence that needs integration.
% 4) dissimilarity uses asymmetric information divergence that requires integration.

We present our \textit{logic embeddings} to address these issues with 
(1) formulation of set logic in terms of first-order logic,
(2) use of well-studied logic,
(3) simple representation with truth bounds, and
(4) fast, symmetric dissimilarity measure.

\textit{Logic embeddings} are a compositional query embedding with inductive bias of real-valued logic, for answering (with uncertainty) existential ($\exists$) FOL multi-hop logical queries over incomplete KGs.
It represents entity subsets with arrays of truth bounds on latent propositions that describe and compress their features and relations.
This allows us to directly use real-valued logic to filter and identify answers.
Truth bounds $[l,u]:0\!\le\!l\!\le\!u\!\le\!1$ from \cite{riegel2020logical}
express uncertainty about truths, stating it can be a value range (e.g. unknown $[0,1]$).
Sum of bound widths model uncertainty, which correlates to answer size.
Now intersection is simply conjunction ($\land$) of bounds to retain only shared propositions, union is disjunction ($\lor$) to retain all propositions, and complement is negation ($\neg$) of bounds to find subsets with opposing propositions.
% It supports first-order logic (FOL) also with disjunction ($\lor$) and existential quantifiers ($\exists$).
% It handles queries in first-order logic (FOL), which includes conjunction ($\land$), disjunction ($\lor$), negation ($\neg$), and existential quantifier ($\exists$) operators.

The novelty of \textit{logic embeddings} is that it 
(a) performs set logic with real-valued logic,
(b) characterizes subsets with truth bounds, and
(c) correlates bounds with uncertainty.
Its benefits are 
(a) improved accuracy with well-studied t-norms,
(b) faster, simplified calculations, and
(c) improved prediction of answer size.

\begin{figure}[!t]
% \vskip -0.1in
\begin{center}
\centerline{\includegraphics[width=0.48\textwidth,trim={0cm 14.1cm 14.7cm 0cm},clip]{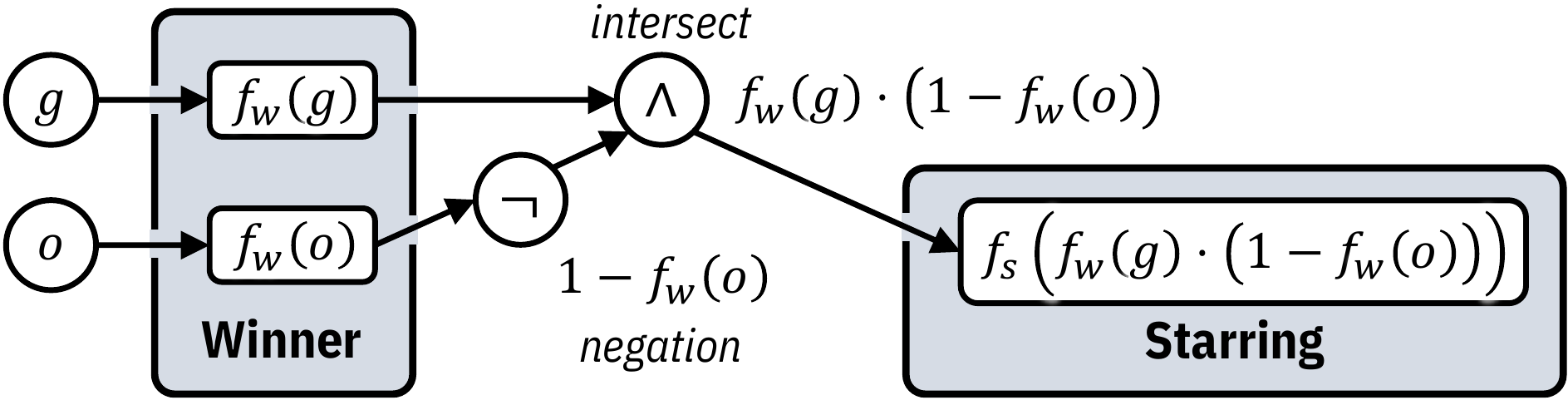}}
\vspace*{-2.5mm}
\caption{Computation graph for Figure~\ref{example1} with product t-norm for intersect. Nodes are truth vectors that identify entity subsets. Embedding query logic reduces to a simple vectorized calculation.}
\label{simple1}
\end{center}
\vskip -0.3in
\end{figure}
Our contributions of \textit{logic embeddings} make several advances to address issues of poor logic and uncertainty modeling, and computational expense of current methods:

\textbf{1. Direct logic.}
Defines Skolem set logic via maximal Skolemisation of first-order logic to embed queries, where proximity of \textit{logic embeddings} directly evaluates logic satisfiability of first-order logic queries.

\textbf{2. Improved logic.}
Performs set intersection as logic conjunction over latent propositions with t-norms, often used for intersection of fuzzy sets. \cite{mostert1957structure} decomposition provide weak, strong, and nilpotent conjunctions, which show higher accuracy than idempotent conjunction via density interpolation of~\cite{ren2020beta}.

\textbf{3. Direct uncertainty.}
Truth bounds naturally model uncertainty and have fast entropy calculation. Both entropy and bounds width show superior correlation to answer size.

\textbf{4. Improved measure.}
Measures dissimilarity with simple $L_1$-norm, which improves training speed and accuracy.

\begin{figure}[!t]
% \vskip -0.1in
\begin{center}
\centerline{\includegraphics[width=0.49\textwidth,trim={0cm 14cm 14.3cm 0cm},clip]{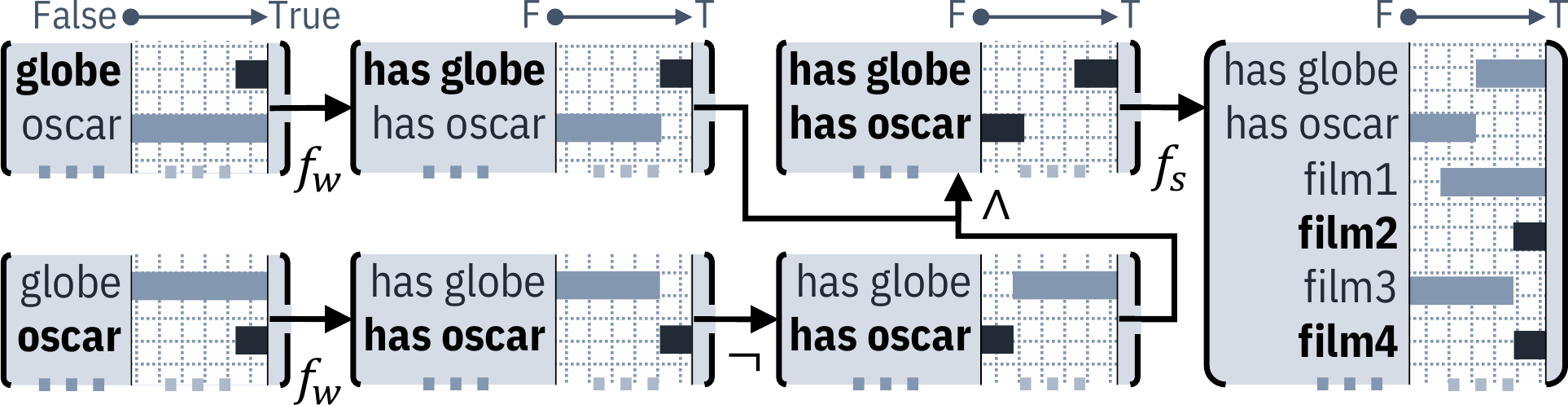}}
\vspace*{-2.5mm}
\caption{Intuition behind truth bounds for Figure~\ref{example1}: propositions identify features, relations substitute propositions, negation flips truths, intersect retains common features. The final query embedding is closer to propositions of film2/4 (answer set) than film1/3.}
\label{truth1}
\end{center}
\vskip -0.3in
\end{figure}

Our contributions also include 
(i) introduction of Skolem set logic in Section 2 to enable lifted inference, 
(ii) definition of \textit{logic embeddings} in Section 3 to enable logic query composition, 
(iii) implementation details of method in Section 4, and 
(iv) detailed evaluation in Section 5 and new cardinality prediction experiment showing benefit of truth bounds.
\section{Querying with Skolem set logic}
We define an existential first-order language $\mathcal{L}$ whose signature contains a set of functional symbols $\mathcal{F}$ and a set of predicate symbols $\mathcal{R}$.
The alphabet of $\mathcal{L}$ includes the logic symbols of conjunction ($\land$), disjunction ($\lor$), negation ($\neg$), and existential quantification ($\exists$).
The semantics of $\mathcal{L}$ interprets the domain of discourse as a family of subsets $\mathcal{C}\subset 2^{\mathcal{V}}$, where $2^{\mathcal{V}}$ denotes the power set of a set of entities $\mathcal{V}$. This allows for lifted inference where variables and terms map to entity subsets, which are single elements in the discourse.

\begin{figure}[!t]
% \vskip -0.1in
\begin{center}
\centerline{\includegraphics[width=0.48\textwidth,trim={0cm 14.8cm 14.7cm 0cm},clip]{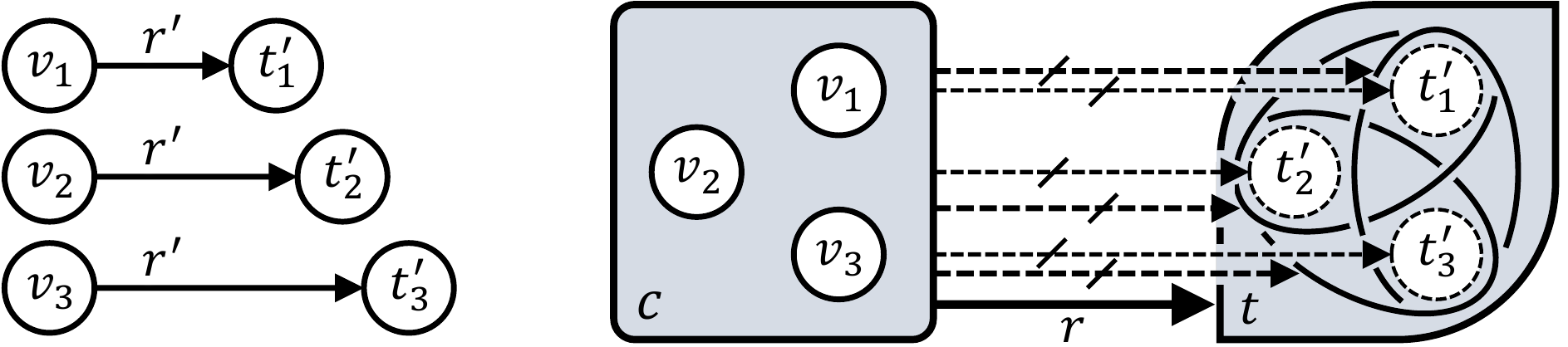}}
\vspace*{-2.5mm}
\caption{A maximum Skolem function relates subset $c$ via $r(c,t)$ to the largest subset $t=\{t_1',t_2',t_3'\}$, and not to smaller subsets, e.g. $\{t_1'\}$ or $\{t_1',t_2'\}$, even if these satisfy $r'(v,t'): v\in c, t'\in t$.}
\label{skolem1}
\end{center}
\vskip -0.3in
\end{figure}

Sentences of $\mathcal{L}$ express relational knowledge, e.g. a binary predicate $r\!\in\!\mathcal{R}$ relates two unordered subsets via function $r\!:\!\mathcal{C}\!\times\!\mathcal{C}\!\mapsto\![0,1]$ that evaluates to a real truth value.
However, the underlying knowledge is predicated under a different signature on domain of discourse $\mathcal{V}$, relating entities via $r'\!:\!\mathcal{V}\!\times\!\mathcal{V}\!\mapsto\! \{\textit{False},\textit{True}\}$.
Subsets $c,t\!\in\!\mathcal{C}$ are related via $r(c,t)$, with the union $t=\cup_{\forall v\in c}t'$ over all underlying propositions $r'(v,t')$ for each entity $v\in c$ (see Figure~\ref{skolem1}).

Existential quantification in $\mathcal{L}$ results in sentences that are always true in the underlying interpretation%
\footnote{Universal quantification is not supported as interpretation in the underlying knowledge is contradictory, since useful relations cannot all be true for both empty and non-empty subsets.}
, because the nullset is present in $\mathcal{C}$.
We introduce maximal quantification via a modified Skolem function to make $\mathcal{L}$ useful.

\begin{figure}[!t]
% \vskip -0.1in
\begin{center}
\centerline{\includegraphics[width=0.48\textwidth,trim={0cm 11.1cm 14.7cm 0cm},clip]{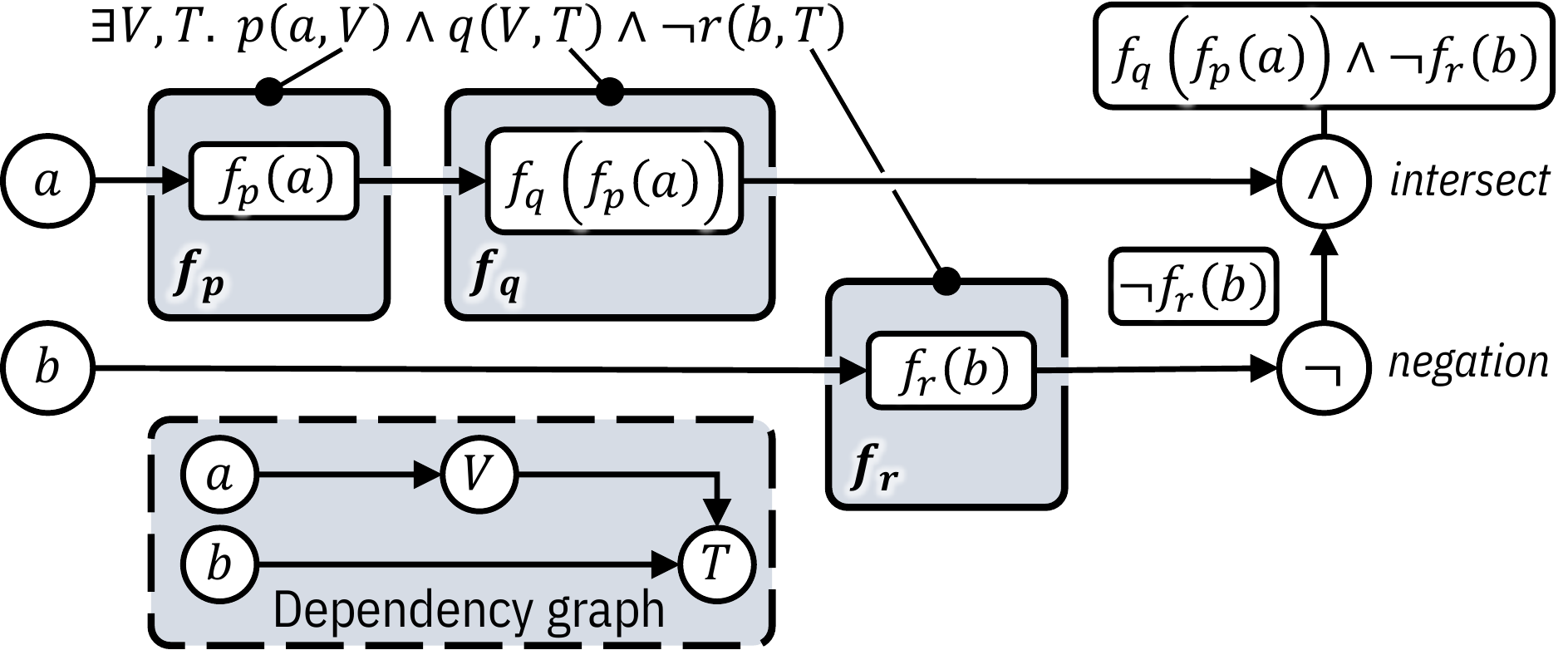}}
\vspace*{-2.5mm}
\caption{Computation and dependency graph for the \textit{pin} query in Table~\ref{tab:queryrep}, we map FOL to Skolem set logic with simple rules.}
\label{form1}
\end{center}
\vskip -0.25in
\end{figure}

\textsl{\textbf{Definition 1}} (Maximal Skolem function).
A maximal Skolem function $f_r\in\mathcal{F}:\mathcal{C}\mapsto\mathcal{C}$ assigns the maximal subset $c$ to an existentially quantified variable $T$ so that $\exists T. r(a,T)\Leftrightarrow r(a,f_r(a))$  (equisatisfiable).
The function receives input element $a\in\mathcal{C}$ and outputs the related subset $c$ over $\mathcal{V}$ with the largest cardinality, so that
$f_r(a)=c:\;|c'|\leq|c|,\;c,\forall c'\in\mathcal{C}$.
$\blacksquare$

Formulae in $\mathcal{L}$ such as $r(a,T)$ can thus convert to sentences $\exists T. r(a,T)$ by quantification over free variables, and the largest satisfying assignment to target variable $T:r(a,T)$ obtained via \textit{maximal Skolemisation} $f_r(a)$ then subsumes all valid groundings in the underlying interpretation over $\mathcal{V}$.
``Relation following" of \cite{cohen2020scalable} is similar where $r(a,T)=\{t'\;|\;\exists v\in a:r'(v,t')\}$.
Normal Skolem functions are different as they only map to single entities.

\textbf{Skolem set logic.}
This is set logic that involves Skolem functions that substitute related subsets.
We introduce notation for \textit{maximal Skolemisation} of sentences in $\mathcal{L}$ that represents set logic on Skolem terms in their underlying interpretation over $\mathcal{V}$. Evaluation in $\mathcal{L}$ of conjunction $\land$, disjunction $\lor$, and negation $\neg$ correspond to intersection, union, and complement in \textit{Skolem set logic}, respectively.%
\footnote{Skolem set logic reuses operators $\land,\lor,\neg$ to signify that it performs direct real-valued logic on truth vectors of its terms. Skolem set logic usages are noted to avoid confusion with FOL.}

\begin{table}[!t]
\vskip -0.1in
\caption{Complex logical query structures of \cite{ren2020beta} in first-order logic and Skolem set logic forms. Notation: intersection (i), union (u), predicate (p), and negation (n).
See Figure~\ref{form1} for example of converting \textit{pin} to Skolem set logic.}
\label{tab:queryrep}
\begin{center}
\begin{small}
\setlength{\tabcolsep}{3pt}
\begin{tabular}{@{}l|l|l@{}} 
\toprule
  & \textbf{First-order logic}& \textbf{Skolem set logic} \\ \midrule
1p & $\exists T.\; p(a,T)$& $f_p(a)$\\
2p & $\exists V,T.\; p(a,V)\land q(V,T)$ & $f_q(f_p(a))$\\
3p & $\exists V,W,T.\; p(a,V)\land q(V,W)$ & $f_r(f_q(f_p(a)))$ \\
& $\quad\land\, r(W,T)$ & \\
\cmidrule{1-3}
2i & $\exists T.\; p(a,T)\land q(b,T)$ & $f_p(a)\land f_q(b)$ \\
3i & $\exists T.\; p(a,T)\land q(b,T)\land r(c,T)$ & $f_p(a)\!\land\!f_q(b)\!\land\!f_r(c)$ \\
pi & $\exists V,T.\; p(a,V)\land q(V,T)\land r(b,T)$ & $f_q(f_p(a))\land f_r(b)$\\
ip & $\exists V,T.\; [p(a,V)\!\land\!q(b,V)]\!\land\!r(V,T)$ & $f_r(f_p(a)\land f_q(b))$\\
\cmidrule{1-3}
2in & $\exists T.\; p(a,T)\land \neg q(b,T)$ & $f_p(a)\land \neg f_q(b)$ \\
3in & $\exists T.\; p(a,T)\land q(b,T)\land\neg r(c,T)$ & $f_p(a)\!\land\! f_q(b)\!\land\!\neg f_r(c)$ \\
pin & $\exists V,T.\; p(a,V)\!\land\!q(V,T)\!\land\!\neg r(b,T)$ & $f_q(f_p(a))\land\neg f_r(b)$\\
pni & $\exists V,T.\; p(a,V)\!\land\!\neg q(V,T)\!\land\! r(b,T)$ & $\neg f_q(f_p(a))\land f_r(b)$\\
inp & $\exists V,T.\; [p(a,V)\land\neg q(b,V)]$ &  $f_r(f_p(a)\land\neg f_q(b))$\\
& $\quad\land\, r(V,T)$ \\
\cmidrule{1-3}
2u & $\exists T.\; p(a,T)\lor q(b,T)$ & $f_p(a)\lor f_q(b)$ \\
up & $\exists V,T.\; [p(a,V)\!\lor\! q(b,V)]\!\land\! r(V,T)$ & $f_r(f_p(a)\lor f_q(b))$\\
\bottomrule
\end{tabular}
\end{small}
\end{center}
\vskip -0.1in
\end{table}

Sentence forms and their \textit{Skolem set logic} representations ($\land$, $\lor$ extends trivially to more inputs) for target variable $T$, given anchor elements $a$, $b$ (assigned subsets) and relation predicates ($r\!,\!\ldots\!,\!z$) include these conversion rules:
\begin{itemize}[leftmargin=1em, nosep, itemsep=1ex]
    \item \textbf{Relation}: $\exists T. r(a,T)$ gives $f_r(a)$.
    \item \textbf{Negation}: $\exists T. \neg r(a,T)$ gives $\neg f_r(a)$.
    \item \textbf{Conjunction}: $\exists T. r(a,T)\land q(b,T)$ gives $f_r(a)\land f_q(b)$.
    \item \textbf{Disjunction}: $\exists T. r(a,T)\lor q(b,T)$ gives $f_r(a)\lor f_q(b)$.
    \item \textbf{Multi-hop}: $\exists T. r(a,V_1)\land q(V_1,V_2)\land\ldots\land z(V_{n-1},T)$ has chain-like relations that give $f_z\cdots (f_q(f_r(a)))$.
\end{itemize}

% with Skolem normal form $p(a,f_p(a))\land q(f_p(a),f_q(f_p(a)))\land\ldots\land z((f_y\circ\cdots\circ f_p)(a), (f_z\circ\cdots\circ f_p)(a))$

\textbf{Query formulae.}
We consider query formulae defined over $\mathcal{V}$ with $m$ anchor entities $a_i\in\mathcal{V}$, a single free target variable $T$, and $n$ bound variables $V_j$.
Query answering assigns the subset $\{t\in\mathcal{V}\}\in\mathcal{C}$ of entities to $T$ that satisfy
\begin{align}
\exists V_1,V_2,\ldots,V_n.\, q(a_1,a_2,\ldots,a_m,V_1,\ldots,V_n,T).
\label{eq:query}
\end{align}
We recast formulae into $\mathcal{C}$ by converting anchor entities to singleton subsets, we then quantify $T$ to \textit{maximally Skolemise} the sentence and derive its \textit{Skolem set logic} term $f(a_1,a_2,\ldots,a_m)$ for $T$ given the anchor entities.

The \textit{dependency graph} of formula~(\ref{eq:query}) consists of vertices $\{a_1,\ldots,a_m,V_1,\ldots,V_n,T\}$ and a directed edge for each vertex pair $(x,y)$ related inside the formula, e.g. via $r(x,y)$.
Queries are valid when the \textit{dependency graph} is a single-sink acyclic graph with anchor entities as source nodes~\cite{hamilton2018embedding}, and an equivalent to the original formula can then be recovered from \textit{Skolem set logic}.
\section{Logic Embeddings}

Our approach for positive inference on queries of form~(\ref{eq:query}) require only \textit{Skolem set logic}, but should also support:
\begin{enumerate}[leftmargin=1.3em, nosep, itemsep=0.7ex]
    \item \textbf{Lifted inference}: Inference over subsets of entities, needing fewer actions than with single-entity inference;
    \item \textbf{Knowledge integration:} Underlying single-entity knowledge over $\mathcal{V}$ integrates over subsets from $\mathcal{C}$;
    \item \textbf{Generalization:} Use of subset similarities to predict absent knowledge with an uncertainty measure.
\end{enumerate}

The powerset $\mathcal{C}$ over entities $\mathcal{V}$ from a typical KB is extremely large, so discrete approaches to achieve above with non-uniform subset representations are likely intractable.

\textbf{Set embeddings}.
We consider \textit{set embeddings} that map $\mathcal{C}$ to a continuous space $\mathcal{M}$, so these images of subsets approximately preserve their relationships from $\mathcal{C}$~\cite{sun2019information}.
It has metric properties such as the volume of subsets, not usually considered by graph embeddings.
% Singleton subsets allow for embedding of single entities.
% \textit{Set embeddings} are uniform, continuous, permutation-invariant, preserve uncertainty and proximity, and are amenable to downstream machine learning tasks:
\textit{Set embeddings} have the following properties:
\begin{itemize}[leftmargin=1em, nosep, itemsep=0.7ex]
    \item \textbf{Uniform}: Enables standard parameterization, simplifies memory structures and related computation;
    \item \textbf{Continuous}: Differentiable, enables optimization;
    \item \textbf{Permutation-invariant}: Subset elements unordered;
    \item \textbf{Uncertainty}: Subset size corresponds to entropy;
    \item \textbf{Proximity}: Relatively preserves subset dissimilarities.
\end{itemize}

% \textit{Set embeddings} consider metric properties such as the volume of subsets, not usually considered by graph embeddings. Singleton subsets allow for embedding of single entities.

\textsl{\textbf{Definition 2}} (Logic embeddings).
\textit{Logic embeddings} are \textit{set embeddings} that characterize subsets with latent propositions, and perform set logic on subsets via logic directly over their latent propositions.
$\blacksquare$

\textit{Logic embeddings} inherit the aforementioned properties and benefits of \textit{set embeddings}, but are also:
\begin{itemize}[leftmargin=1em, nosep, itemsep=0.7ex]
    \item \textbf{Logical}: Logic over truth values in embeddings performs set logic, and proximity correlates with satisfiability;
    \item \textbf{Contextual}: Latent propositions integrate select knowledge depending on the subset;
    \item \textbf{Open-world}: Accepts and integrates unknown or partially known knowledge and inferences.
\end{itemize}

\textit{Logic embeddings} also share query embedding advantages of efficient answering, generalization, and full logic support:
\begin{itemize}[leftmargin=1.3em, nosep, itemsep=0.7ex]
    \item \textbf{Fast querying}: Obtains answers closest to query embedding in sublinear time, unlike subgraph matching with exponential time in query size~\cite{dalvi2007efficient}.
    \item \textbf{General querying}: Generalizes to unseen query forms.
    \item \textbf{Implicit prediction}: Implicitly imputes missing relations, and avoids exhaustive link prediction~\cite{de2008logical} subgraph matching requires and scales poorly on.
    \item \textbf{Natural modeling}: Supports intuitive set intersection, unlike point embeddings~\cite{hamilton2018embedding}.
    \item \textbf{Uncertainty}: Models answer size with embedding entropy~\cite{ren2020beta} or truth bounds (ours).
    \item \textbf{Fundamental support}: Handles negation (and disjunction via De Morgan's law), unlike box embeddings where complements are not closed regions (and union resorts to disjunctive normal form)~\cite{ren2020query2box}.
\end{itemize}

\textbf{Latent propositions.}
A \textit{logic embedding} keeps truth values with associate distribution $\mathbf{p}_X\in\mathcal{M}$ on latent propositions of features and properties that characterize and distinguish subset $X\subset\mathcal{V}$.
Subset entities $x\in X$ may share a similar relation $r(x,Y)$ to a particular subset $Y$, where latent propositions can integrate such identifying relations.
\textit{Logic embeddings} need to be contextual given the limited embedding capacity, as only some relations may be relevant to define a particular subset.

\textbf{Uncertainty.}
We represent volume in embedding space with lower and upper bounds $[l,u]$ on truth values, to express uncertainty and allow correlation of embedding entropy of a subset with its cardinality.
We use truth bounds of \cite{riegel2020logical} that admit the open-world assumption and have probabilistic semantics to interpret known ($l\leq u$), unknown ($[0,1]$) and contradictory states ($l>u$, not considered here).

The \textit{logic embedding} for $X\subset\mathcal{V}$ is an $n$-tuple $\mathbf{S}_X=([l_i,u_i]: l_i,u_i\in[0,1])_{i=1}^n$ of lower and upper bound pairs ($l_i\leq u_i$) that represents an $n$-tuple $\mathbf{p}_X=(P_i)_{i=1}^n$ of uniform distributions $P_i=U(l_i,u_i)$, which omits contradiction $l_i>u_i$.
The chain rule for differential entropy $H(\mathbf{p}_X)=H(P_1,\ldots,P_n)$ of the embedding distribution applies and gives an upper-bound in terms of components $H(P_i)=\log(u_i-l_i)$, where
$$H(P_1,\ldots,P_n)=\sum_{i=1}^n H(P_i|P_1,\ldots,P_{i-1})\leq\sum_{i=1}^n H(P_i).
$$
\textsl{\textbf{Condition 1}} (Uncertainty axiom).
Set embeddings should (approx.) satisfy $\forall X\in\mathcal{C}$: entropy $H(\mathbf{p}_X)$ is a monotonically increasing function of $H(U_X)$, where $U_X$ is a uniform distribution over elements of $X$~\cite{sun2019information}.
$\blacksquare$

We measure adherence to the uncertainty axiom with correlation between subset size $|X|$ and entropy upper-bound $\sum_{i=1}^n H(P_i)$ or total truth interval width $\sum_{i=1}^n(u_i-l_i)$, and by predicting $|X|$ from $\mathbf{h}_X=[H(P_i)]_{i=1}^n$.

\textbf{Proximity.}
Subsets with high overlap should embed close by, whereas little to no overlap should result in relatively distant embeddings.
We now review the proximity axiom.

\textbf{\textsl{Condition 2}} (Proximity axiom).
$\forall(X,X')\in\mathcal{C}^2$: $D(\mathbf{p}_{X}\|\mathbf{p}_{X'})$ should positively correlate with $D(U_{X}\|U_{X'})$, given information divergence $D$~\cite{sun2019information}.
$\blacksquare$

Relative entropy is an important divergence where the family of $f$-divergences $D_f(p|q)=\int p(x)f(q(x)/p(x))dx$ typically include $\log(p(x))$ or $p(x)^{-1}$ terms over finite support $x\in\mathcal{X}$.
Uniform distributions $U(l,u)$ in \textit{logic embeddings} may not cover a $[0,1]$ support, and may result in undefined divergence.
Therefore, we measure dissimilarity $D(\mathbf{S}_X,\mathbf{S}_{X'})\in[0,1]$ between \textit{logic embeddings} of subsets $(X,X')$ with the expected mean of $L_1$-norms of truth bounds, where
\begin{align}
    D(\mathbf{S}_X,\mathbf{S}_{X'})=\sum_{i=1}^n \frac{\|l_i-l_i'\|+\|u_i-u_i'\|}{2n}.
\end{align}
\textbf{\textsl{Condition 3}} (Satisfiability axiom).
Substitution instance $q(X')$ of first-order logic formula $q(T)$ has satisfiability $1-D(\mathbf{S}_X,\mathbf{S}_{X'})$, where target variable $T$ has answer $X$.
$\blacksquare$

Query $q(T)$ is true for answer $X$, since $1-D(\mathbf{S}_X,\mathbf{S}_{X})=1$,
but candidate satisfiability $q(X')$ can reduce to minimum 0 (false), depending on the dissimilarity between $X'$ and $X$.

\textbf{Set logic.}
Conjunction and disjunction of latent propositions of subsets perform their intersection and union, respectively, unlike \textit{information-geometric} set embeddings that interpolate distributions. 
De Morgan's law replaces disjunction $a\lor b$ with conjunction and negations $\neg(\neg a\!\land\!\neg b)$.
Involute negation of  $\mathbf{S}\!=\!\neg(\neg\mathbf{S})\!=\!([l_i,u_i])_{i=1}^n$ describes complement%
\footnote{Truth bounds can share intervals with their negations, therefore complements can share latent propositions, which supports the diverse partitioning that complex queries require.}
$\neg\mathbf{S}=\left(\left[1\!-\!u_i,1\!-\!l_i\right]\right)_{i=1}^n$.
We use continuous t-norm $\top:[0,1]^k\mapsto [0,1]$
to perform generalized conjunction for real-valued logic, and calculate
$\mathbf{S}'=\bigwedge_{j=1}^k\mathbf{S}_j$
as
\begin{align}
\mathbf{S}'=\left(\left[\top(l_i^{(1)},\ldots,l_i^{(k)}), \top(u_i^{(1)},\ldots,u_i^{(k)})\right]\right)_{i=1}^n.
\end{align}
\cite{mostert1957structure} decompose any continuous t-norm into Archimedean t-norms, namely minimum/G\"odel $\top_{\text{min}}(\mathbf{t})=\min(t_1,\ldots,t_k)$, product $\top_{\text{prod}}(\mathbf{t})=\prod_{j=1}^kt_j$, and \L ukasiewicz $\top_{\text{luk}}(\mathbf{t})=\max(0,1-\sum_{j=1}^k(1-t_j))$, which we evaluate separately to consider all prime aspects.

\textbf{Contextual.}
Limited capacity requires intersection to reintegrate latent propositions contextually via a weighted t-norm: 1) continuous function%
\footnote{We use non-monotonic smoothmin ($\alpha\!=\!-10$) for weighted minimum t-norm and set $l'\!=\!u'\!=\!(l+u)/2$ when $l\!>\!u$.} 
$\top(\mathbf{w},\mathbf{t})$ of weights $\mathbf{w}$ and truths $\mathbf{t}$; 2) behaves equal to unweighted case if weights are 1; and 3) $w_j=0$ removes input $j$, and weights are in $[0,1]$.
\begin{align}
    \top_{\text{min}}(\mathbf{w},\mathbf{t})&=\textstyle\sum_{j=1}^kt_jw_je^{\alpha t_j}\left/\textstyle\sum_{j=1}^kw_je^{\alpha t_j}\right.\\
    \top_{\text{prod}}(\mathbf{w},\mathbf{t})&=\textstyle\prod_{j=1}^kt_j^{w_j}\\
    \top_{\text{luk}}(\mathbf{w},\mathbf{t})&=\max\left(0,1-\textstyle\sum_{j=1}^kw_j(1-t_j)\right)
\end{align}
Weight $w_j^{(v)}$ for $(l_j^{(v)},u_j^{(v)})$, the $j^{\text{th}}$ truth bounds in input $v$, depends on bounds of all conjunction inputs via attention, starting with function $g$ as
\begin{align}
g_j^{(v)}=g(l_1^{(v)},\ldots,l_n^{(v)},u_1^{(v)},\ldots,u_n^{(v)}).
\end{align}
Softargmax over all the conjunction inputs yields a score
$s_j^{(v)}=\exp(g_j^{(v)})/\sum_{h=1}^k\exp(g_j^{(h)})$
which normalizes after
$w_j^{(v)}=s_j^{(v)}/\max(s_j^{(1)},\ldots,s_j^{(k)})$
to ensure max weight 1.
\section{Implementation}

\textbf{Query embedding.}
We calculate a \textit{logic embedding} for a single-sink acyclic query with \textit{Skolem set logic} over anchor entities.
We keep vectors $\{\mathbf{r}\in\mathbb{R}^d\}$ for relation embeddings, and $\{\mathbf{x}\in[0,1]^{2d}\}$ for \textit{logic embeddings} of all entities, where $\mathbf{x}=[l_1,\ldots,l_d,u_1,\ldots,u_d]$.
To measure the ``cost" of modeling uncertainty by tracking bounds we also test point truth embeddings ($l=u$), where $\mathbf{x}=[t_1,\ldots,t_{2d}]$.

We parameterize our Skolem function $f_r(\mathbf{x})=f(\mathbf{r},\mathbf{x})$
with
$\mathbf{F}_1\in\mathbb{R}^{3d\times h}$,
$\mathbf{F}_2\in\mathbb{R}^{h\times h}$, and
$\mathbf{F}_3\in\mathbb{R}^{h\times 2d}$
to relate $\mathbf{x}$ to
$\mathbf{y}=[\mathbf{y}_l,\mathbf{y}_l+\mathbf{y}_u'(1-\mathbf{y}_l)]$, where
$[\mathbf{y}_l,\mathbf{y}_u']=f'(\mathbf{r},\mathbf{x})=\sigma\left(\max\left(0,\max\left(0,[\mathbf{r},\mathbf{x}]\mathbf{F}_1\right)\mathbf{F}_2\right)\mathbf{F}_3\right)$
activates sigmoid.

Set logic has
attention that uses $g(\mathbf{x})=\max\left(0,\mathbf{x}\mathbf{G}_1\right)\mathbf{G}_2$
with parameter matrices
$\mathbf{G}_1\in\mathbb{R}^{2d\times 2d}$ and
$\mathbf{G}_2\in\mathbb{R}^{2d\times d}$.

\textbf{Cardinality prediction.}
We predict%
\footnote{1:1 train:test, $\rho=10^3$, 250 epochs, Adam opt. ($lr=10^{-4}$).}
the cardinality $|X|$ of subset $X$ from the entropy vector $\mathbf{h}_X$ of its \textit{logic embedding} with
    $\rho\cdot\sigma\left(\max\left(0,\max\left(0,\mathbf{h}_X\mathbf{H}_1\right)\mathbf{H}_2\right)\mathbf{H}_3\right)$ scaled by $\rho$, where
$\mathbf{H}_1\in\mathbb{R}^{d\times \frac{d}{4}}$,
$\mathbf{H}_2\in\mathbb{R}^{\frac{d}{4}\times \frac{d}{16}}$, and
$\mathbf{H}_3\in\mathbb{R}^{\frac{d}{16}\times 1}$.%

\textbf{Query answering.}
Our objective is to embed a query $\mathbf{q}$ relatively close to its answers $\{\mathbf{y}\}$ and far from negative samples $\{\mathbf{z}\}$. We train model parameters%
\footnote{Hyperparameters include $d=400$, $h=1600$, $\gamma=0.375$, $k=128$ random negative samples, $512$ batch size, $450$k epochs, Adam optimizer ($lr=10^{-4}$). Pytorch on 1x NVIDIA Tesla V100.}
of 1) entity \textit{logic embeddings}, 2) relation embeddings, 3) Skolem function, and 4) t-norm attention, to minimize query answering loss
\begin{align}\label{eq:loss}
    -\log\sigma(\gamma-D(\mathbf{y},\mathbf{q}))-\sum_{j=1}^k\frac{1}{k}\log\sigma(D(\mathbf{z}_j,\mathbf{q})-\gamma).
\end{align}
\section{Experiments}
We primarily compare against Beta embeddings (\betae \cite{ren2020beta}) that also support arbitrary FOL queries and negation, where our \textit{logic embeddings} (\logice with \L ukasiewicz t-norm) show improved 1) generalization, 2) reasoning, 3) uncertainty modeling, and 4) training speed.

\textbf{Datasets.}
We use two complex logical query datasets from: \qb with 9 query structures \cite{ren2020query2box}, and \betae that adds 5 for negation \cite{ren2020beta}.
They generate random queries separately over three standard KGs with official train/valid/test splits, namely FB15k~\cite{bordes2013translating}, FB15k-237~\cite{toutanova2015observed}, NELL995~\cite{xiong2017deeppath}.
Table~\ref{tab:queryrep} shows the first-order logic and Skolem set logic forms of the 14 query templates.

We separately follow the evaluation procedures of above \qb and \betae datasets.%
\footnote{https://github.com/francoisluus/KGReasoning}
Training omits ip/pi/2u/up (Table~\ref{tab:queryrep}) to test handling of unseen query forms.
Negation is challenging with 10x less queries than conjunctive ones.%
\footnote{Please see appendix for statistics of datasets.}

\textbf{Generalization.}
Queries have at least one link prediction task to test generalization, where withheld data contain goal answers.
We measure Hits@k and mean reciprocal rank (MRR) of these non-trivial answers that do not appear in train/valid data.
Table~\ref{tab:mrr_betae_small} tests both disjunctive normal form (DNF) and De Morgan's form (DM) for unions (2u/up), but we only report DNF elsewhere as it outperforms DM.

\logice with bounds generalizes better than \betae, \qb, and \gqe \cite{hamilton2018embedding} on almost all query forms in Table~\ref{tab:mrr_betae_small}, and further improves with point truths.%
\footnote{\label{fn:sub1}Please see appendix for full results on FB15k and NELL.}
\logice also answers negation queries more accurately than \betae for most query forms in Table~\ref{tab:mrr_neg_betae_small}.
CQD-Beam does not handle negation nor uncertainty and is expensive as it grounds candidate entities explicitly, yet \logice generalizes better and more efficiently in Table~\ref{tab:emql}(a) on FB15k-237 and NELL.

\begin{table}[!t]
\caption{Test MRR results (\%, higher better) of \logice and \betae on answering queries with negation (\betae dataset).\footnoteref{fn:sub1}}
\label{tab:mrr_neg_betae_small}
\setlength{\tabcolsep}{4.1pt}
\vskip 0.15in
\begin{center}
\begin{small}
\begin{tabular}{l|ccccc|c|cc}
\toprule
 & \multicolumn{6}{c|}{\textbf{FB15k-237}} & \textbf{FB15k} & \textbf{NELL}\\
\hline
\textbf{Model} & \textbf{2in} & \textbf{3in} & \textbf{inp} & \textbf{pin} & \textbf{pni} & \textbf{avg}  & \textbf{avg}  & \textbf{avg} \\
\midrule
\logicept & \textbf{4.9} & \textbf{8.2} & \textbf{7.7} & \textbf{3.6} & \textbf{3.5} & \textbf{5.6} & \textbf{12.5} & 6.2 \\
\logiceb & 4.9 & 8.0 & 7.3 & 3.6 & 3.5 & 5.5 & 11.7 & \textbf{6.3} \\
\betae & 5.1 & 7.9 & 7.4 & 3.6 & 3.4 & 5.4 & 11.8 & 5.9 \\
\bottomrule
\end{tabular}
\end{small}
\end{center}
\vskip -0.3in
\end{table}

\textbf{Reasoning.}
Logical entailment on queries without missing links tests how faithful deductive reasoning is.
We thus train on all splits and measure entailment accuracy in Table~\ref{tab:emql}(b).
\logice on average reasons more faithfully than \betae, \qb, and \gqe baselines on all datasets.

EmQL is a query embedding that specifically optimises faithful reasoning \cite{sun2020faithful}, and thus outperforms all other methods in Table~\ref{tab:emql}(b).
However, EmQL without its sketch method has worse faithfulness than \logice with point truths for FB15k and NELL, also EmQL does not support negation nor models uncertainty like \logice.

\textbf{Compare logics.}
\logice can intersect via minimum, product, or \L ukasiewicz t-norms, which perform similarly ($\pm$1\%) in Table~\ref{tab:mrr_att}, while all outperform \betae.
\L ukasiewicz provides superior uncertainty modeling, so is the default choice for \logice.
Attention via weighted t-norm improves \logice accuracy (+9.6\%), where one hypothesis is better use of limited embedding capacity through learning weighted combinations of latent propositions.

\begin{table}[!t]
\caption{Validation MRR averages (\%, higher better) for \logice with various t-norms and \betae on training queries (\betae datasets), where i, n, and p are all query forms containing intersection, negation, or relation components, respectively. }
\label{tab:mrr_att}
\setlength{\tabcolsep}{3.7pt}
\vskip 0.15in
\begin{center}
\begin{small}
    \begin{tabular}{l|rrrr|rrrr|r}
\toprule
\textbf{Model} & \multicolumn{4}{c}{\textbf{FB15k-237}} & \multicolumn{4}{|c|}{\textbf{NELL995}} & \\
% \cmidrule(l{2pt}r{2pt}){2-5}
% \cmidrule(l{2pt}r{2pt}){6-9}
& \multicolumn{1}{c}{i} & \multicolumn{1}{c}{n} & \multicolumn{1}{c}{p} & \multicolumn{1}{c|}{all} & \multicolumn{1}{c}{i} & \multicolumn{1}{c}{n} & \multicolumn{1}{c}{p} & \multicolumn{1}{c|}{all} & \multicolumn{1}{c}{avg}\\
\midrule
\multicolumn{10}{l}{\textit{No bounds with attention} (\logice)} \\
\cmidrule{1-4}
luk & 14.4 & 5.0 & 13.2 & 15.8 & \textbf{18.8} & 6.5 & \textbf{19.5} & 21.5 & 18.7 \\
min & 14.4 & 5.0 & 13.2 & 15.7 & 18.6 & 6.6 & 19.4 & 21.5 & 18.6 \\
prod & \textbf{14.4} & \textbf{5.0} & \textbf{13.3} & \textbf{15.8} & 18.7 & \textbf{6.6} & 19.4 & \textbf{21.5} & \textbf{18.7} \\
\midrule
\multicolumn{10}{l}{\textit{Bounds with attention} (\logiceb)} \\
\cmidrule{1-5}
luk & 13.9 & 5.0 & 12.9 & 15.3 & \textbf{18.7} & 6.5 & 19.0 & 21.3 & 18.3 \\
% luk$\sim$ & 13.8 & 4.8 & 12.9 & 15.2 & 18.6 & 6.5 & 18.9 & 21.3 & 18.2 \\
min & 13.9 & 5.0 & 12.7 & 15.2 & 18.5 & 6.6 & 19.0 & 21.3 & 18.2 \\
prod & \textbf{14.0} & \textbf{5.0} & \textbf{13.0} & \textbf{15.4} & 18.6 & \textbf{6.7} & \textbf{19.1} & \textbf{21.3} & \textbf{18.4} \\
\betae & 13.7 & 4.8 & 12.6 & 15.0 & 17.5 & 6.1 & 17.0 & 19.6 & 17.3 \\
\midrule
\multicolumn{10}{l}{\textit{Bounds with no attention}}\\
\cmidrule{1-6}
luk & 12.7 & 5.1 & 11.7 & 14.1 & 16.1 & 6.8 & 16.7 & 18.9 & 16.5 \\
min & \textbf{13.0} & \textbf{5.4} & 11.8 & \textbf{14.4} & 16.4 & 7.1 & 16.8 & 19.1 & 16.8 \\
prod & 12.9 & 5.3 & \textbf{11.9} & 14.3 & \textbf{16.5} & \textbf{7.1} & \textbf{17.0} & \textbf{19.2} & \textbf{16.8} \\
\betae & 11.6 & 5.0 & 11.7 & 13.3 & 15.4 & 5.7 & 14.8 & 17.5 & 15.4 \\
\bottomrule
\end{tabular}
\end{small}
\end{center}
\vskip -0.3in
\end{table}
However, \betae improves by avg. +12.3\% with a similar attention mechanism so has greater dependence on it, possibly because it devises intersection as interpolation of densities, whereas \logice uses established real-valued logic via t-norms. In particular, the \betae intersect is idempotent while \logice offers weak and strong conjunctions of which \L ukasiewicz offers nilpotency.

\begin{table*}[!htpb]
\caption{Test MRR results (\%, higher better) of \logice, \betae, \qb and \gqe on answering EPFO ($\exists$, $\wedge$, $\vee$) queries (\betae dataset).\footnoteref{fn:sub1}}
\label{tab:mrr_betae_small}
\vskip 0.15in
\begin{center}
\begin{small}
\begin{tabular}{l|ccc|cc|cc|cc|cc|c|cc}
\toprule
 & \multicolumn{12}{c|}{\textbf{Generalization on FB15k-237}} & \textbf{FB15k} & \textbf{NELL}\\
\hline
\multirow{2}{*}{\textbf{Model}} & \multirow{2}{*}{\textbf{1p}} & \multirow{2}{*}{\textbf{2p}} & \multirow{2}{*}{\textbf{3p}} & \multirow{2}{*}{\textbf{2i}} & \multirow{2}{*}{\textbf{3i}} & \multirow{2}{*}{\textbf{pi}} & \multirow{2}{*}{\textbf{ip}} & \multicolumn{2}{c|}{\textbf{2u}} & \multicolumn{2}{c|}{\textbf{up}} & \multirow{2}{*}{\textbf{avg}}  & \multirow{2}{*}{\textbf{avg}}  & \multirow{2}{*}{\textbf{avg}} \\
 & & & & & & & & \textbf{DNF} & \textbf{DM} & \textbf{DNF} & \textbf{DM} & \\
\midrule
\logicept &  \textbf{41.3} & \textbf{11.8} & \textbf{10.4} & \textbf{31.4} & \textbf{43.9} & \textbf{23.8} & \textbf{14.0} & \textbf{13.4} & \textbf{13.1} & \textbf{10.2} & 9.8 & \textbf{22.3} & \textbf{44.1} & \textbf{28.6}\\
\logiceb & 40.5 & 11.4 & 10.1 & 29.8 & 42.2 & 22.4 & 13.4 & 13.0 & 12.9 & 9.8 & 9.6 & 21.4 & 40.8 & 28.0 \\
\betae & 39.0 & {10.9} & {10.0} & 28.8 & {42.5} & {22.4} & {12.6} & {12.4} & 11.1 & {9.7} & \textbf{9.9} & {20.9} & 41.6 & 24.6 \\
\qb & {40.6} & 9.4 & 6.8 & {29.5} & 42.3 & 21.2 & {12.6} & 11.3 & - & 7.6 & - & 20.1 & 38.0 & 22.9\\
\gqe & 35.0 & 7.2 & 5.3 & 23.3 & 34.6 & 16.5 & 10.7 & 8.2 & - & 5.7 & - & 16.3 & 28.0 & 18.6 \\
\bottomrule
\end{tabular}
\end{small}
\end{center}
\vskip -0.1in
\end{table*}
\begin{table*}[!htbp]
\caption{Hits@3 results (higher better) on the \qb datasets testing (a) generalization and (b) reasoning faithfulness.\footnoteref{fn:sub1}}
\label{tab:emql}
\vskip 0.15in
\begin{center}
\begin{small}
\begin{tabular}{l|ccccc|cccc|c|cc}
\toprule
   \multicolumn{11}{c|}{(a) \textbf{Generalization on FB15k-237} (\qb datasets)} & \textbf{FB15k} & \textbf{NELL}\\
\hline
%  Model & 1p   & 2p   & 3p   & 2i   & 3i   & ip   & pi   & 2u   & up  & avg & avg & avg \\
 \textbf{Model} & \textbf{1p}   & \textbf{2p}   & \textbf{3p}   & \textbf{2i}   & \textbf{3i}   & \textbf{ip}   & \textbf{pi}   & \textbf{2u}   & \textbf{up}  & \textbf{avg} & \textbf{avg} & \textbf{avg} \\
   \midrule
 \logicept & 46.1 & 28.6 & 24.8 & 34.8 & 46.5 & 12.0 & 23.7 & {27.7} & 21.1 & 29.5 & {54.9} & 39.3 \\
 \logiceb & 45.0 & 26.6 & 23.0 & 32.0 & 44.1 & 11.1 & 22.1 & 25.5 & 20.4 & 27.7 & 50.3 & 38.6 \\
 EmQL & 37.7 & \textbf{34.9} & \textbf{34.3} & \textbf{44.3} & \textbf{49.4} & \textbf{40.8} & \textbf{42.3} & 8.7  & \textbf{28.2} & \textbf{35.8} & {49.5} & \textbf{46.8}\\
%  ~~$-$ sketch & 43.1 & 34.6 & 33.7 & 41.0 & 45.5 & 36.7 & 37.2 & 15.3 & \textbf{32.5} & 35.5 & 48.6 & \textbf{46.8}\\
 CQD-Beam & \textbf{51.2} & 28.8 & 22.1 & 35.2 & 45.7 & 12.9 & 24.9 & \textbf{28.4} & 12.1 & 29.0 & \textbf{68.0} & 37.5\\
 \betae & 43.1 & 25.3 & 22.3 & 31.3 & 44.6 & 10.2 & 22.3 & 26.6 & 18.0 & 27.1 & 51.4 & 33.8 \\
 \qb & 46.7 & 24.0   & 18.6 & 32.4 & 45.3 & 10.8 & 20.5 & {23.9} & 19.3 & 26.8 & 48.4 & 30.6 \\
 %  ~~$+d$=2000            & 37.2  & 20.7  & 19.4  & 22.6  & 37.1  & 9.7  & 16.8  & 20.0  & 17.8 & 22.4 & 34.5 & 23.4 \\
 \gqe & 40.5 & 21.3 & 15.5 & 29.8 & 41.1 & 8.5  & 18.2 & 16.9 & 16.3 & 23.1 & 38.7 & 24.8 \\
\midrule
   \multicolumn{11}{c|}{(b) \textbf{Entailment on FB15k-237} (\qb datasets)} & \textbf{FB15k} & \textbf{NELL}\\
\midrule
 \logicept & 81.5 & 54.2 & 46.0 & 58.1 & 67.1 & 28.5 & 44.0 & 66.6 & 40.8 & 54.1 & 65.5 & 85.3 \\
 \logiceb & 73.7 & 46.4 & 38.9 & 49.8 & 61.5 & 22.0 & 37.2 & 54.6 & 35.1 & 46.6 & 58.4 & 80.1 \\
 EmQL & \textbf{100.0}  & \textbf{99.5} & \textbf{94.7} & \textbf{92.2} & \textbf{88.8} & \textbf{91.5} & \textbf{93.0} & \textbf{94.7} & \textbf{93.7} & \textbf{94.2} & \textbf{91.4} & \textbf{98.8} \\ 
~~$-$ sketch & 89.3 & 55.7 & 39.9 & 62.9 & 63.9 & 51.9 & 54.7 & 53.8 & 44.7 & 57.4 & 55.5 & 82.5 \\
 \betae & 77.9 & 52.6 & 44.5 & 59.0 & 67.8 & 23.5 & 42.2 & 63.7 & 35.1 & 51.8 & 60.6 & 80.2 \\
 \qb & 58.5 & 34.3 & 28.1 & 44.7 & 62.1 & 11.7 & 23.9 & 40.5 & 22.0 & 36.2 & 43.7 & 51.1 \\
% ~~+$d$=2000            & 50.7 & 30.1 & 26.1 & 34.8 & 55.2 & 11.4 & 20.6 & 32.8 & 21.5 & 31.5 & 38.3 & 43.7\\
 \gqe & 56.4 & 30.1 & 24.5 & 35.9 & 51.2 & 13.0 & 25.1 & 25.8 & 22.0 & 31.6 & 43.7 & 49.8\\
 \bottomrule
\end{tabular}
\end{small}
\end{center}
\vskip -0.1in
\end{table*}
\begin{table*}[!htpb]
\caption{Spearman's rank correlation and Pearson's correlation coefficient (higher better) between learned embedding (diff. entropy and truth interval width for \logice, diff. entropy for \betae, $L_1$ box size for \qb) and the number of answers of queries (\betae dataset).\footnoteref{fn:sub1}}
\label{tab:spearman_small}
\vskip 0.15in
\begin{center}
\begin{small}
\begin{tabular}{l|ccc|cc|cc|ccccc|c|cc}
\toprule
 & \multicolumn{13}{c|}{\textbf{Spearman's rank correlation on FB15k-237}} & \textbf{FB15k} & \textbf{NELL}\\
\hline
\textbf{Model} & \textbf{1p} & \textbf{2p} & \textbf{3p} & \textbf{2i} & \textbf{3i} & \textbf{pi} & \textbf{ip} & \textbf{2in} & \textbf{3in} & \textbf{inp} & \textbf{pin} & \textbf{pni} & \textbf{avg}  & \textbf{avg}  & \textbf{avg} \\
\midrule
Entropy & \textbf{0.65} & \textbf{0.67} & \textbf{0.72} & 0.61 & 0.51 & \textbf{0.57} & \textbf{0.60} & \textbf{0.69} & 0.54 & \textbf{0.62} & \textbf{0.61} & \textbf{0.67} & \textbf{0.62} & \textbf{0.58} & \textbf{0.61} \\
Interval & 0.61 & 0.58 & 0.58 & \textbf{0.64} & \textbf{0.64} & 0.54 & 0.49 & 0.58 & 0.50 & 0.41 & 0.49 & 0.60 & 0.56 & 0.51 & 0.53 \\
\betae & 0.40 & 0.50 & 0.57 & 0.60 & 0.52 & 0.54 & 0.44 & 0.69 & \textbf{0.58} & 0.51 & 0.47 & 0.67 & 0.54 & 0.49 & 0.55\\
\qb & 0.18 & 0.23 & 0.27 & 0.35 & 0.44 & 0.36 & 0.20 & - & - & - & - & - & - & - & -\\
\midrule
& \multicolumn{13}{c|}{\textbf{Pearson correlation coef. on FB15k-237}} & \textbf{FB15k} & \textbf{NELL}\\
\midrule
Entropy & \textbf{0.33} & \textbf{0.53} & \textbf{0.61} & \textbf{0.45} & 0.37 & \textbf{0.37} & \textbf{0.47} & \textbf{0.58} & \textbf{0.44} & \textbf{0.52} & \textbf{0.49} & \textbf{0.57} & \textbf{0.48} & \textbf{0.46} & \textbf{0.52} \\
%  & \betae & 0.196 & 0.358 & 0.417 & 0.357 & 0.293 & 0.325 & 0.337 & 0.431 & 0.380 & 0.366 & 0.338 & 0.428 & 0.352 \\ % own results
\betae & 0.23 & 0.37 & 0.45 & 0.36 & 0.31 & 0.32 & 0.33 & 0.46 & 0.41 & 0.39 & 0.36 & 0.48 & 0.37 & 0.36 & 0.4\\
\qb & 0.02 & 0.19 & 0.26 & 0.37 & \textbf{0.49} & 0.34 & 0.20 & - & - & - & - & - & - & - & - \\
\bottomrule
\end{tabular}
\end{small}
\end{center}
\vskip -0.1in
\end{table*}
\begin{table*}[!htpb]
\caption{Answer size prediction mean absolute error (\%, lower better) with embedding entropy components for \logice and \betae, and box size components for \qb (\betae dataset).\footnoteref{fn:sub1}}
\label{tab:sizepredict}
\vskip 0.15in
\begin{center}
\begin{small}
\begin{tabular}{l|ccc|cc|cc|ccccc|c|cc}
\toprule
& \multicolumn{13}{c|}{\textbf{Answer size prediction error on FB15k-237}} & \textbf{FB15k} & \textbf{NELL}\\
\hline
\textbf{Model} & \textbf{1p} & \textbf{2p} & \textbf{3p} & \textbf{2i} & \textbf{3i} & \textbf{pi} & \textbf{ip} & \textbf{2in} & \textbf{3in} & \textbf{inp} & \textbf{pin} & \textbf{pni} & \textbf{avg}  & \textbf{avg}  & \textbf{avg} \\
\midrule
\logice & \textbf{78} & \textbf{83} & \textbf{86} & \textbf{82} & \textbf{94} & \textbf{89} & \textbf{86} & \textbf{81} & \textbf{79} & \textbf{81} & \textbf{81} & \textbf{81} & \textbf{83} & \textbf{87} & \textbf{80} \\
\betae & 111 & 96 & 97 & 97 & 97 & 95 & 97 & 97 & 95 & 97 & 97 & 98 & 98 & 95 & 95 \\
\qb & 191 & 101 & 100 & 310 & 780 & 263 & 103 & - & - & - & - & - & - & - & - \\
\bottomrule
\end{tabular}
\end{small}
\end{center}
\vskip -0.1in
\end{table*}

\textbf{Uncertainty modeling.}  
Correlation between differential entropy $\sum_{i=1}^n H(P_i)$ (upper-bound) and answer size (uncertainty, number of entities) is significantly higher in \logice than \betae using both Spearman's rank correlation and Pearson's correlation coefficient in Table~\ref{tab:spearman_small}. Both significantly outperform uncertainty of \qb with $L_1$ box size.

Total truth interval width $\sum_{i=1}^n(u_i-l_i)$ of \logice correlates better to answer size in most cases than \betae, and offers direct use of the probabilistic semantics of truth bounds to simplify uncertainty modeling.
Note that minimizing query answering loss in Eq.~(\ref{eq:loss}) does not directly optimize answer cardinality, so \logice naturally models uncertainty only as by-product of learning to answer.

\textbf{Cardinality prediction.}
Above evaluation aggregates entropy, but element-wise entropies
$\mathbf{h}_X=[H(P_i)]_{i=1}^n$
contain more information that we use for explicit answer size prediction.
We train a regression classifier to map provided uncertainties $\mathbf{h}_X$ to answer size $|X|$, and measure mean absolute error $\|s-|X|\|/|X|$ of size prediction $s$.

Table~\ref{tab:sizepredict} shows reduced cardinality prediction error of 83\% with \logice, compared to avg. 96\% with \betae, indicating more informative uncertainties with \logice.
However, these errors are still quite large, possibly because the main training objective does not directly optimize uncertainties for cardinality prediction.

\begin{figure}[!t]
\vskip 0.2in
\begin{center}
\centerline{
\begin{subfigure}[t]{0.17\textwidth}%
    \centering%
    \captionsetup{font=small}%
    \includegraphics[width=\textwidth,trim={0.18cm 0cm 0cm 0cm},clip]{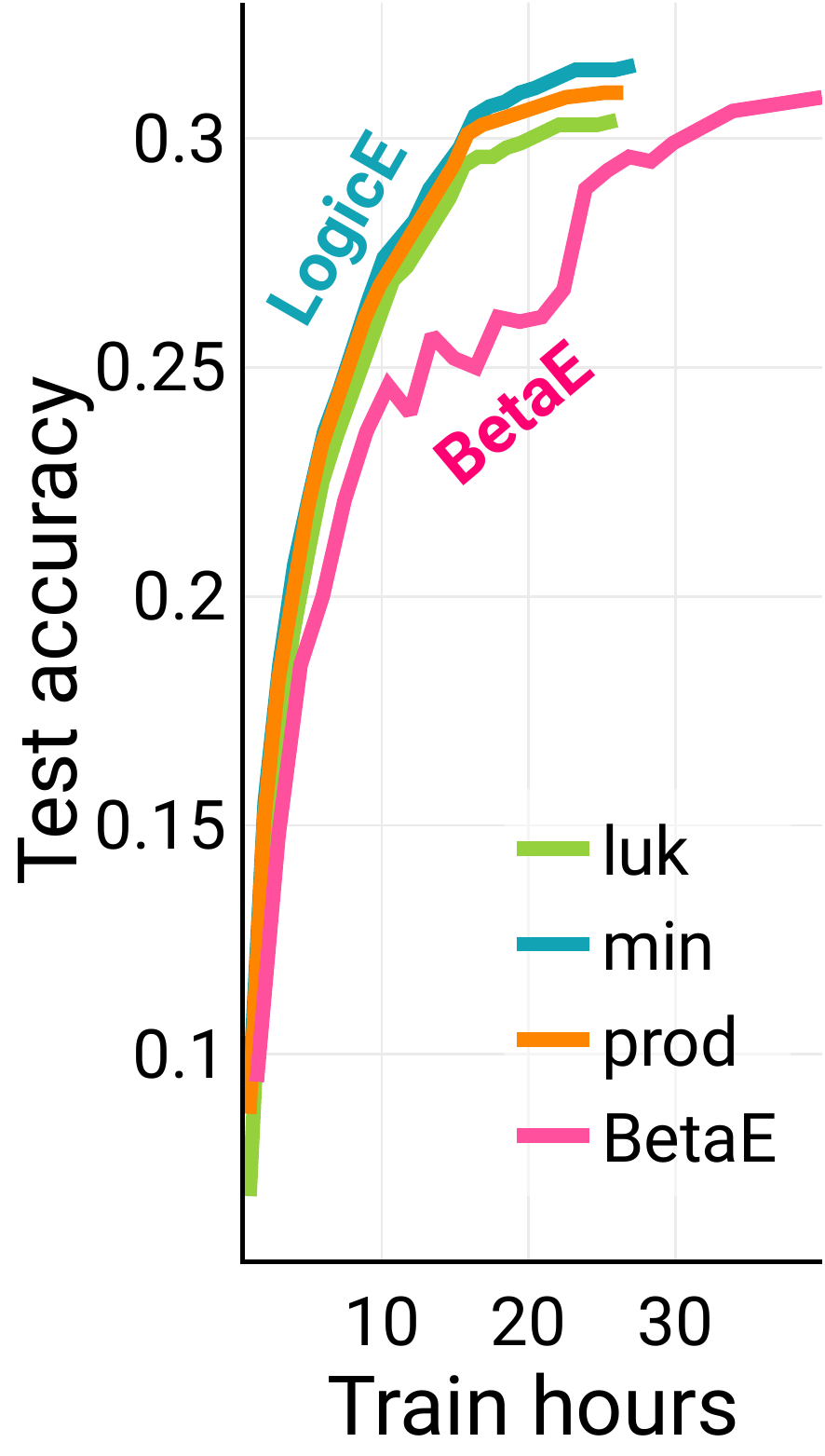}%
    \caption{FB15k}
    \label{fig:traintime0}%
\end{subfigure}
\begin{subfigure}[t]{.153\textwidth}%
    \centering%
    \captionsetup{font=small}%
    \includegraphics[width=\textwidth,trim={0cm 0cm 0cm 0cm},clip]{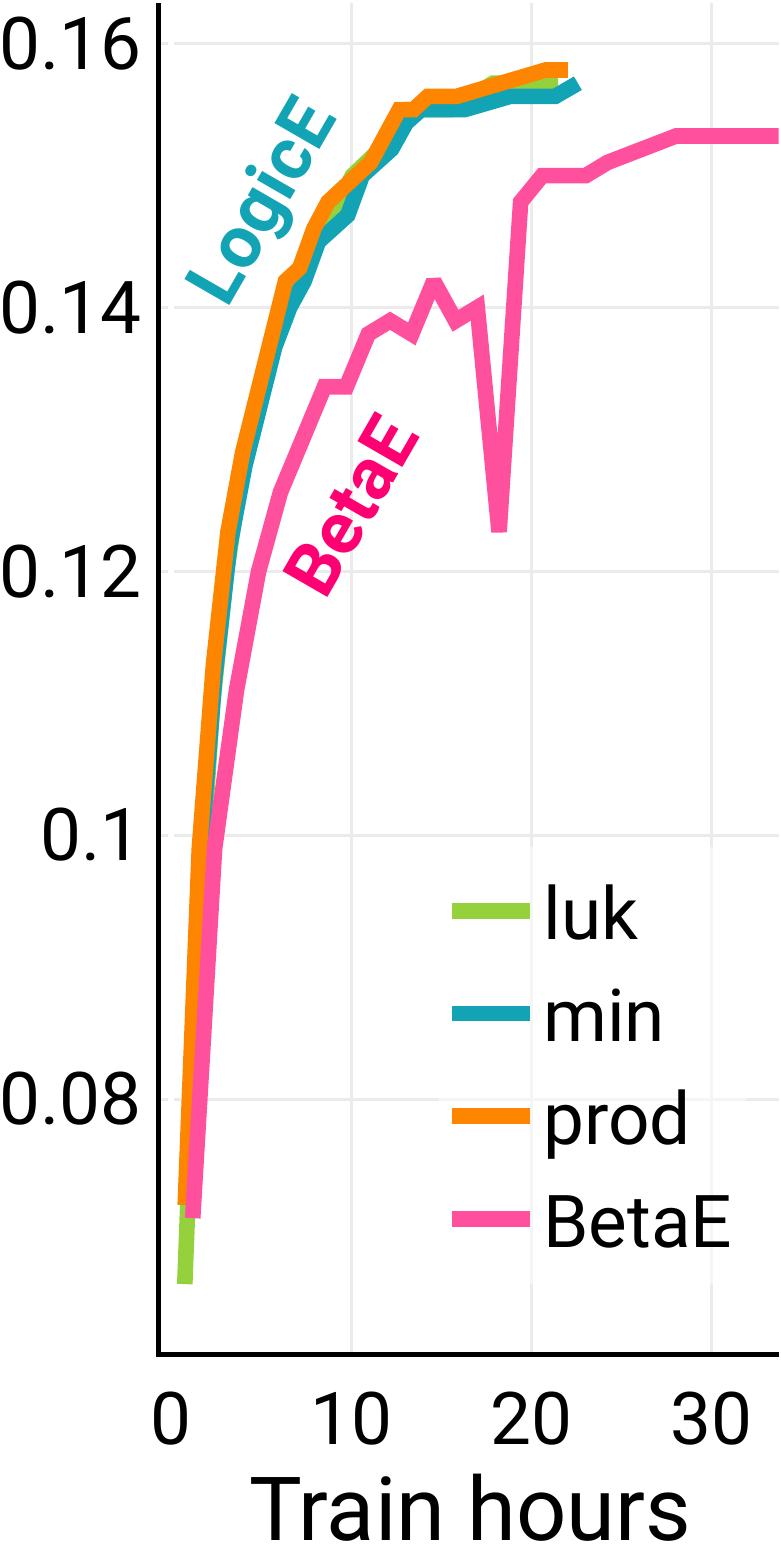}%
    \caption{FB15k-237}
    \label{fig:traintime1}%
\end{subfigure}
\begin{subfigure}[t]{.153\textwidth}%
    \centering%
    \captionsetup{font=small}%
    \includegraphics[width=\textwidth,trim={0cm 0cm 0cm 0cm},clip]{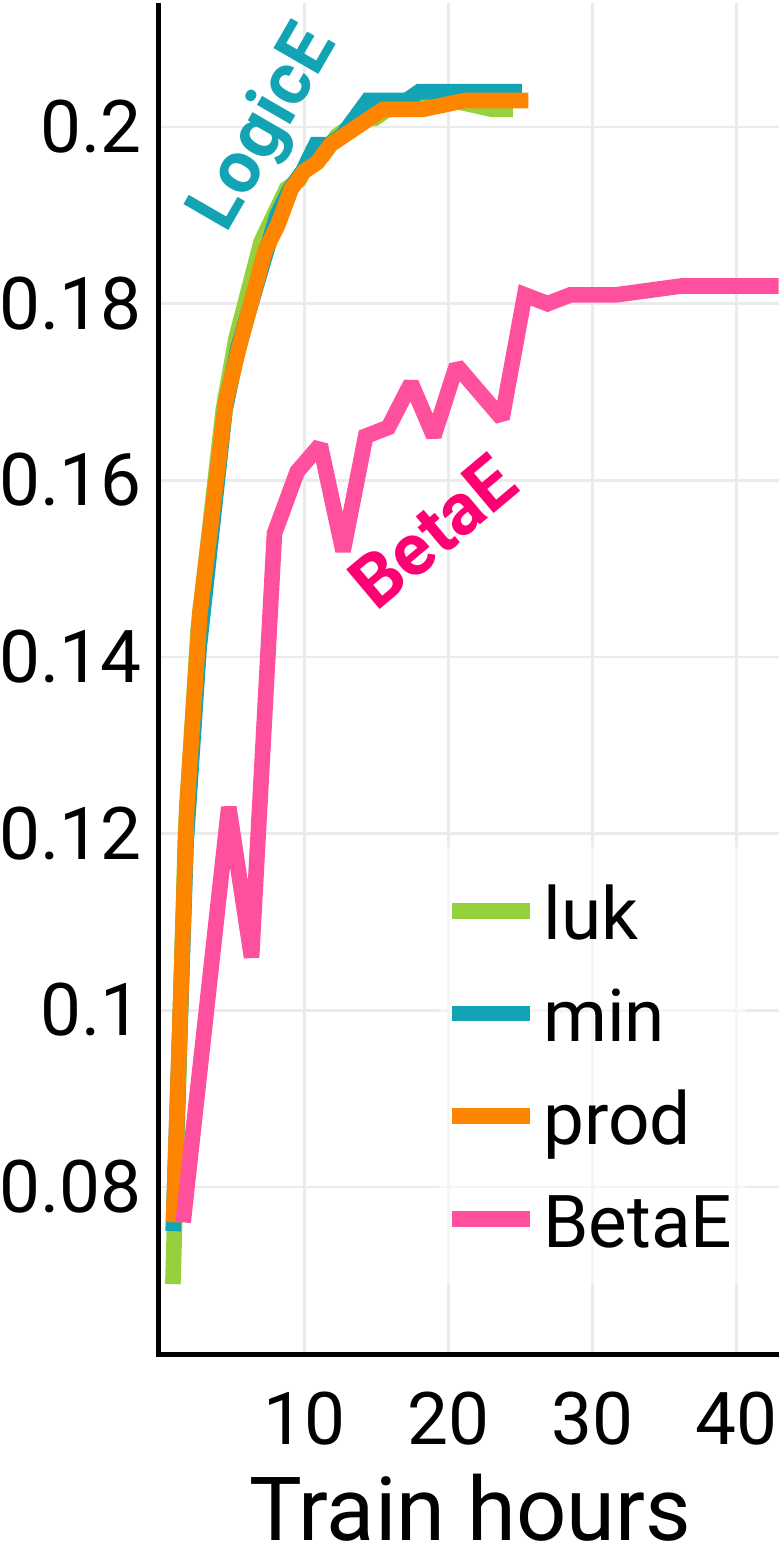}%
    \caption{NELL995}
    \label{fig:traintime2}%
\end{subfigure}
}
\caption{Test accuracy vs. training time comparison (V100).}
\label{fig:traintime}
\end{center}
\vskip -0.3in
\end{figure}

\textbf{Training speed.}
\betae uses the Beta distribution with no closed form that requires integration to compute entropy and dissimilarity. In contrast, \logice uses simple truth bounds with fast entropy and dissimilarity calculations.
Figure~\ref{fig:traintime} shows that \logice with any t-norm trains 2-3x faster than \betae, with the exact same compute resources, optimizer and learning rate.
The \logice training curve also appears smooth and monotonic, compared to disrupted learning progress with \betae.

\section{Related work}
In addition to the overview of query embeddings in the Introduction, we also relate our work to 1) tensorized logic, 2) querying with t-norms, and 3) lifted inference.%
\footnote{Please see appendix for an extended related work section.}

\textbf{Tensorized logic.}
Distributed representations like embeddings can enable generalization and efficient inference that symbolic logic lacks.
Logic tensors of \cite{grefenstette2013towards} express truths as specific tensors and map entities to one-hot vectors with full-rank matrices, but only memorize facts.
Matrix factorization reduces one-hot vectors to low dimensions to enable generalization and efficiency while optimizing logic constraints, but can scale exponentially in the number of query variables \cite{rocktaschel2015injecting}.

Logic Tensor Networks learn a real vector per entity and even Skolem functions that map to entity features, but has weak inductive bias as it needs to learn predicates to perform logic \cite{serafini2016logic}.
In contrast, \textit{logic embeddings} support uncertainty and only has to learn Skolem functions to express knowledge and generalize, with direct logic on latent truths and logic satisfiability via distance.

\textbf{Querying with t-norms.}
Triangular norms allow for differentiable composition of scores, often in the context of expensive search.
\cite{guo2016jointly} jointly embed KGs and logic rules via t-norm of scores, but only for simple rules.
\cite{arakelyan2020complex} combine scores from a pretrained link predictor via t-norms repeatedly to search for an answer while tracking intermediaries, whereas \textit{logic embeddings} perform vectorized t-norm to directly embed answers.

\textbf{Lifted inference.}
Many probabilistic inference algorithms accept first-order specifications, but perform inference on a mostly propositional level by instantiating first-order constructs~\cite{friedman1999learning, richardson2006markov}.
In contrast, lifted inference operates directly on first-order representations, manipulating not only individuals but also groups of individuals, which has the potential to significantly speed up inference~\cite{de20071}.
% Variable elimination of non-observed non-query variables is the basis of several lifted inference algorithms~\cite{poole2003first, de20071}, and strongly resembles the \textit{lifting lemma} which simulates ground resolution because it is complete and then lifts the resolution proof to the first-order world~\cite{chang2014symbolic}.
% Theorem proving produces a potentially unbounded number of resolutions on grounded representations by performing unification and resolution on clauses with free variables, operating directly on the first-order representation.
% However, r
% Reasoning over incomplete knowledge requires generalization where particular facts about some individuals could apply with uncertainty to a similar group, thus predicting missing facts.
We need to reason about entities we know about, as well as those we know exist but with unknown properties. 
% \textit{Logic embeddings} perform lifted inference as it does not have to reason separately for each entity.
\textit{Logic embeddings} perform a type of lifted inference as it does not have to ground out the theory or reason separately for each entity, but can perform logic inference directly on subsets.

\section{Conclusion}
Embedding complex logic queries close to answers is efficient but presents several difficulties in set theoretic modeling of uncertainty and full existential FOL, where Euclidean geometry and probability density approaches suffer deficiency and computational expense.
Our \textit{logic embeddings} overcome these difficulties by converting set logic into direct real-valued logic.
% , to directly compose an FOL query with logic on latent truth values that identify subsets.
We execute FOL queries logically, and not through Venn diagram models like other embeddings, yet we achieve efficiency of lifted inference over subsets.
% We use truth bounds with fast and simple dissimilarity and entropy measures, and can measure logic satisfiability directly.

Main limitations include the need for more training data than search-based methods, although we have strong inductive bias of t-norm logic to reduce sample size dependence.
Future work will consider negation attending to applied relations of its input to benefit from context like the t-norms.


\bibliography{refs}
\bibliographystyle{icml2021}

\newpage
\appendix

\section{Extended Related Work}

\textbf{Link prediction.}
Reasoning over knowledge bases is fundamental to Artificial Intelligence, but still challenging since most knowledge graphs (KGs) 
such as DBpedia \cite{bizer2009dbpedia}, Freebase \cite{bollacker2008freebase}, and NELL \cite{carlson2010toward} 
are often large and incomplete.
Answering complex queries is an important use of KGs, but missing facts makes queries unanswerable under normal inference.
KG embeddings are popular for predicting facts, learning entities as vectors and relations between them as functions in vector space, like translation \cite{bordes2013translating} or rotation \cite{sun2019rotate}.

Link prediction uncovers similar behavior of entities, and semantic similarity between relations (e.g. birthplace predicts nationality).
Path queries involve multi-hop reasoning (e.g. country of birthcity of person), where compositional learning embeds queries close to answer entities with fast (sublinear) neighbor search \cite{guu2015traversing}.
In contrast, sequential path search grows exponentially in the number of hops, and requires approaches like reinforcement learning \cite{das2017go} or beam search \cite{arakelyan2020complex} that have to explicitly track intermediate entities.

\textbf{Path-based methods.}
A simple approach to complex query answering represents first-order logical queries as a directed graph corresponding to the reasoning path to be followed. Such path-based methods  are  characterized by carrying out a sub-graph matching strategy in their pursuit for solving complex queries. However, they fail to deal with queries with missing relations and cannot scale to large KGs as the complexity of sub-graph matching grows exponentially in the query size. Several works aim at addressing the former by imputing missing relations \cite{guu2015traversing,hong2018page}, leading into a denser KG with high computational demand. 

\textbf{Query embeddings.}
Recent approaches aim to address the two issues by learning embeddings of the query such that entities that answer the query are close to the embeddings of the query and answers can be found by fast nearest neighbor searches. Such approaches implicitly impute missing relations and also lead to faster querying compared to subgraph matching. Here, logical queries as well as KG entities are embedded into a lower-dimensional vector space as geometric shapes such as points \cite{hamilton2018embedding}, boxes \cite{ren2020query2box} and distributions with bounded support \cite{ren2020beta}.

Compared to point-based embeddings, boxes and distributions naturally model sets of entities they enclose, with set operations on those sets corresponding to logical operations (e.g., set intersection corresponds to the conjunction operator), and thus iteratively executing set operations results in logical reasoning. Furthermore, box and distribution-based embeddings allow handling the uncertainty over the queries. 

Majority of early embedding based approaches are limited to a subset of first-order logic consisting of existential quantification and conjunctions, with a few recent papers supporting the so-called existential positive first-order (EPFO) queries \cite{ren2020query2box,arakelyan2020complex} that additionally include disjunctions. The work by \cite{ren2020beta} is the first to handle the full set of first-order logic including negation. 

Recent path-based approaches utilize knowledge graph embeddings to learn to tractably traverse the graph in the embedding space \cite{lou2020neural} or using pre-trained black boxes for link prediction \cite{arakelyan2020complex}. Neural Subgraph Matching \cite{lou2020neural} uses order embeddings to embed the query and KG graphs into a lower-dimensional space, and efficiently performs subgraph matching directly in the embedding space. This has the potential to impute missing relations and has been shown to be orders of magnitude faster than standard sub-graph matching approaches on subgraph matching benchmarks. However, it has not been applied to complex query answering problems. \cite{arakelyan2020complex} use a pre-trained, black-box, neural link predictor to reduce sample complexity and scale to larger KGs, and was shown to be effective on EPFO queries, but does not support the full set of first-order logic queries.

\textbf{Lifted inference.}
Many probabilistic inference algorithms accept first-order specifications, but perform inference on a mostly propositional level by instantiating first-order constructs~\cite{friedman1999learning, richardson2006markov}.
In contrast, lifted inference operates directly on first-order representations, manipulating not only individuals but also groups of individuals, which has the potential to significantly speed up inference~\cite{de20071}.

Variable elimination of non-observed non-query variables is the basis of several lifted inference algorithms~\cite{poole2003first, de20071}, and strongly resembles the \textit{lifting lemma} which simulates ground resolution because it is complete and then lifts the resolution proof to the first-order world~\cite{chang2014symbolic}.

Theorem proving produces a potentially unbounded number of resolutions on grounded representations by performing unification and resolution on clauses with free variables, operating directly on the first-order representation.
However, reasoning over incomplete knowledge requires generalization where particular facts about some individuals could apply with uncertainty to a similar group, thus predicting missing facts.

We need to reason about entities we know about, as well as entities we know exist but which have unknown properties. 
Logic embeddings perform a type of lifted inference as it does not have to ground out the theory or reason separately for each entity, but can perform logic inference directly with compact representations of smooth sets of entities.

\section{Query forms}
Table~\ref{tab:queryrep0} gives logic query forms from \cite{ren2020beta} with corresponding Skolem normal form and Skolem set logic.
Note that Skolem normal form uses conventional set logic symbols inside predicates, to avoid confusion with logic symbols composing predicates.
Subsequent Skolem set logic resumes use of logic symbols, as the actual operations are logical over vectors of truth values.

\section{Dataset statistics}
Table~\ref{tab:kgstat} gives the relation and entity counts, as well as the train/valid/test split edge counts for the three datasets used.
Table~\ref{tab:q2bstat} gives the number of generated queries for the \qb query datasets, and Table~\ref{tab:qstat} gives the query counts for the \betae query datasets.

\section{Additional Results}
Table~\ref{tab:mrr_neg_betae} compares test MRR results for \logice and \betae on different datasets.
Table~\ref{tab:mrr_betae} compares test MRR results for queries without negation.
Table~\ref{tab:spearman} provides full details of Spearman's rank correlation for all three KGs.
Table~\ref{tab:pearson} gives full numbers on Pearson's correlation coefficient for all three KGs.
Table~\ref{tab:sizepredict0} gives answer size prediction mean absolute error for all three KGs.
Table~\ref{tab:emql_full} provides generalization and entailment scores for all three KGs.
\begin{table}[!htpb]
\caption{Test MRR results (\%) of \logice and \betae on answering queries with negation.}
\label{tab:mrr_neg_betae}
\setlength{\tabcolsep}{4.1pt}
\vskip 0.15in
\begin{center}
\begin{small}
\begin{tabular}{l|l|ccccc|c}
\toprule
\textbf{Dataset} & \textbf{Model} & \textbf{2in} & \textbf{3in} & \textbf{inp} & \textbf{pin} & \textbf{pni} & \textbf{avg} \\
\midrule
\multirow{1}{*}{FB15k} & \logicept & \textbf{15.1} & 14.2 & \textbf{12.5} & \textbf{7.1} & \textbf{13.4} & \textbf{12.5} \\
 & \logiceb & 14.0 & 13.4 & 11.9 & 6.6 & 12.4 & 11.7 \\
 & \betae & 14.3 & \textbf{14.7} & 11.5 & 6.5 & 12.4 & 11.8 \\
\midrule
\multirow{1}{*}{FB15k-237} & \logicept & \textbf{4.9} & \textbf{8.2} & \textbf{7.7} & \textbf{3.6} & \textbf{3.5} & \textbf{5.6} \\
 & \logiceb & 4.9 & 8.0 & 7.3 & 3.6 & 3.5 & 5.5 \\
 & \betae & 5.1 & 7.9 & 7.4 & 3.6 & 3.4 & 5.4 \\
\midrule
\multirow{1}{*}{NELL995} & \logicept & \textbf{5.3} & 7.5 & 11.1 & 3.3 & 3.8 & 6.2 \\
 & \logiceb & 5.3 & \textbf{7.8} & \textbf{11.1} & \textbf{3.3} & \textbf{3.8} & \textbf{6.3} \\
 & \betae & 5.1 & 7.8 & 10.0 & 3.1 & 3.5 & 5.9 \\
\bottomrule
\end{tabular}
\end{small}
\end{center}
\vskip -0.1in
\end{table}
\begin{table*}[!htbp]
\caption{Complex logical query structures of \cite{ren2020beta} in first-order logic and Skolem set logic form.}
\label{tab:queryrep0}
\vskip 0.15in
\begin{center}
\begin{small}
\begin{tabular}{llll} 
\toprule
  & \textbf{First-order logic} & \textbf{Skolem normal form} & \textbf{Skolem set logic} \\ \midrule
1p & $\exists T.\; p(a,T)$ & $p(a,f_p(a))$ & $f_p(a)$\\
2p & $\exists V,T.\; p(a,V)\land q(V,T)$ & $p(a,f_p(a))\land q(f_p(a),f_q(f_p(a)))$ & $f_q(f_p(a))$\\
3p & $\exists V,W,T.\; p(a,V)\land q(V,W)\land r(W,T)$ & $p(a,f_p(a))\land q(f_p(a),f_q(f_p(a)))$ & $f_r(f_q(f_p(a)))$ \\
 & & $\quad\land\, r(f_q(f_p(a)),f_r(f_q(f_p(a))))$ & \\
\cmidrule{1-3}
2i & $\exists T.\; p(a,T)\land q(b,T)$ & $p(a,f_p(a))\land q(b,f_q(b))$ & $f_p(a)\land f_q(b)$ \\
3i & $\exists T.\; p(a,T)\land q(b,T)\land r(c,T)$ & $p(a,f_p(a))\land q(b,f_q(b))\land r(c,f_r(c))$ & $f_p(a)\land f_q(b)\land f_r(c)$ \\
pi & $\exists V,T.\; p(a,V)\land q(V,T)\land r(b,T)$ & $p(a,f_p(a))\land q(f_p(a),f_q(f_p(a)))\land r(b,f_r(b))$ & $f_q(f_p(a))\land f_r(b)$\\
ip & $\exists V,T.\; [p(a,V)\land q(b,V)]\land r(V,T)$ & $[p(a,f_p(a))\land q(b,f_q(b))]$ & $f_r(f_p(a)\land f_q(b))$\\
 & & $\quad\land\, r(f_p(a)\cap f_q(b),f_r(f_p(a)\cap f_q(b)))$ & \\\cmidrule{1-3}
2in & $\exists T.\; p(a,T)\land \neg q(b,T)$ & $p(a,f_p(a))\land \neg q(b,f_q(b))$ & $f_p(a)\land \neg f_q(b)$ \\
3in & $\exists T.\; p(a,T)\land q(b,T)\land\neg r(c,T)$ & $p(a,f_p(a))\land q(b,f_q(b))\land\neg r(c,f_r(c))$ & $f_p(a)\land f_q(b)\land\neg f_r(c)$ \\
pin & $\exists V,T.\; p(a,V)\land q(V,T)\land\neg r(b,T)$ & $p(a,f_p(a))\land q(f_p(a),f_q(f_p(a)))\land\neg r(b,f_r(b))$ & $f_q(f_p(a))\land\neg f_r(b)$\\
pni & $\exists V,T.\; p(a,V)\land\neg q(V,T)\land r(b,T)$ & $p(a,f_p(a))\land\neg q(f_p(a),f_q(f_p(a)))\land r(b,f_r(b))$ & $\neg f_q(f_p(a))\land f_r(b)$\\
inp & $\exists V,T.\; [p(a,V)\land\neg q(b,V)]\land r(V,T)$ & $[p(a,f_p(a))\land\neg q(b,f_q(b))]$ & $f_r(f_p(a)\land\neg f_q(b))$\\
 & & $\quad\land\, r(f_p(a)\cap\neg f_q(b),f_r(f_p(a)\cap\neg f_q(b)))$ & \\ \cmidrule{1-3}
2u & $\exists T.\; p(a,T)\lor q(b,T)$ & $p(a,f_p(a))\lor q(b,f_q(b))$ & $f_p(a)\lor f_q(b)$ \\
up & $\exists V,T.\; [p(a,V)\lor q(b,V)]\land r(V,T)$ & $[p(a,f_p(a))\lor q(b,f_q(b))]$ & $f_r(f_p(a)\lor f_q(b))$\\
 & & $\quad\land\, r(f_p(a)\cup f_q(b),f_r(f_p(a)\cup f_q(b)))$ & \\
\bottomrule
\end{tabular}
\end{small}
\end{center}
\vskip -0.1in
\end{table*}

\begin{table*}[!htbp]
\caption{Dataset statistics according to \cite{ren2020beta} with training, validation and test edge splits.}
\label{tab:kgstat}
\vskip 0.15in
\begin{center}
\begin{small}
\begin{tabular}{l|cc|ccc|c}
\toprule
\textbf{Dataset} & \textbf{Relations} & \textbf{Entities} & \textbf{Training Edges} & \textbf{Validation Edges} & \textbf{Test Edges} & \textbf{Total Edges}\\
\midrule
FB15k & 1,345 & 14,951 & 483,142 & 50,000 & 59,071 & 592,213\\
FB15k-237 & 237  & 14,505& 272,115 & 17,526 & 20,438 & 310,079\\
NELL995 & 200  & 63,361& 114,213 & 14,324 & 14,267 & 142,804\\
\bottomrule
\end{tabular}
\end{small}
\end{center}
\vskip -0.1in
\end{table*}

\begin{table*}[!htbp]
\caption{Number of queries in \qb dataset generated for different query structures (see \cite{ren2020query2box}).}
\label{tab:q2bstat}
\vskip 0.15in
\begin{center}
\begin{small}
\begin{tabular}{l|cc|cc|cc}
\toprule
\textbf{} & \multicolumn{2}{c|}{\textbf{Training}} & \multicolumn{2}{c|}{\textbf{Validation}} & \multicolumn{2}{c}{\textbf{Test}} \\
\hline
\textbf{Dataset}     & 1p               & others           & 1p              & others          & 1p               & others          \\ \midrule
FB15k     & 273,710          & 273,710          & 59,097          & 8,000           & 67,016           & 8,000           \\
FB15k-237 & 149,689          & 149,689          & 20,101          & 5,000           & 22,812           & 5,000           \\
NELL995     & 107,982          & 107,982          & 16,927          & 4,000           & 17,034           & 4,000           \\ 
\bottomrule
\end{tabular}
\end{small}
\end{center}
\vskip -0.1in
\end{table*}
\begin{table*}[!htbp]
\caption{Number of queries in \betae dataset generated for different query structures (see \cite{ren2020beta}).}
\label{tab:qstat}
\vskip 0.15in
\begin{center}
\begin{small}
\begin{tabular}{l|cc|cc|cc}
\toprule
\textbf{} & \multicolumn{2}{c|}{\textbf{Training}} & \multicolumn{2}{c|}{\textbf{Validation}} & \multicolumn{2}{c}{\textbf{Test}} \\
\hline
\textbf{Dataset}     & 1p/2p/3p/2i/3i               & 2in/3in/inp/pin/pni           & 1p              & others          & 1p               & others          \\ \midrule
FB15k     & 273,710          & 27,371          & 59,097          & 8,000           & 67,016           & 8,000           \\
FB15k-237 & 149,689          & 14,968          & 20,101          & 5,000           & 22,812           & 5,000           \\
NELL995     & 107,982          & 10,798          & 16,927          & 4,000           & 17,034           & 4,000           \\ 
\bottomrule
\end{tabular}
\end{small}
\end{center}
\vskip -0.1in
\end{table*}
\begin{table*}[!htpb]
\caption{Test MRR results (\%) of \logice, \betae, \qb and \gqe on answering EPFO ($\exists$, $\wedge$, $\vee$) queries (\betae data).}
\label{tab:mrr_betae}
\vskip 0.15in
\begin{center}
\begin{small}
\begin{tabular}{l|l|ccc|cc|cc|cc|cc|c}
\toprule
\multirow{2}{*}{\textbf{Dataset}} & \multirow{2}{*}{\textbf{Model}} & \multirow{2}{*}{\textbf{1p}} & \multirow{2}{*}{\textbf{2p}} & \multirow{2}{*}{\textbf{3p}} & \multirow{2}{*}{\textbf{2i}} & \multirow{2}{*}{\textbf{3i}} & \multirow{2}{*}{\textbf{pi}} & \multirow{2}{*}{\textbf{ip}} & \multicolumn{2}{c|}{\textbf{2u}} & \multicolumn{2}{c|}{\textbf{up}} & \multirow{2}{*}{\textbf{avg}} \\
 & & & & & & & & & \textbf{DNF} & \textbf{DM} & \textbf{DNF} & \textbf{DM} & \\ \midrule
\multirow{1}{*}{FB15k}
 & \logicept & \textbf{72.3} & \textbf{29.8} & \textbf{26.2} & \textbf{56.1} & 66.3 & 42.7 & \textbf{32.6} & \textbf{43.4} & \textbf{37.1} & \textbf{27.5} & \textbf{26.7} & \textbf{44.1} \\
 & \logiceb & 67.8 & 27.1 & 24.6 & 51.4 & 61.3 & 41.3 & 30.1 & 38.1 & 33.7 & 25.1 & 23.8 & 40.8 \\
 & \betae & 65.1 & {25.7} & {24.7} & {55.8} & \textbf{66.5} & \textbf{43.9} & {28.1} & {40.1} & 25.0 & {25.2} & 25.4 & {41.6} \\
 & \qb & {68.0} & 21.0 & 14.2 & 55.1 & {66.5} & 39.4 & 26.1 & 35.1 & - & 16.7 & - & 38.0 \\
 & \gqe & 54.6 & 15.3 & 10.8 & 39.7 & 51.4 & 27.6 & 19.1 & 22.1 & - & 11.6 & - & 28.0 \\ \midrule
\multirow{1}{*}{FB15k-237} 
 & \logicept &  \textbf{41.3} & \textbf{11.8} & \textbf{10.4} & \textbf{31.4} & \textbf{43.9} & \textbf{23.8} & \textbf{14.0} & \textbf{13.4} & \textbf{13.1} & \textbf{10.2} & 9.8 & \textbf{22.3} \\
 & \logiceb & 40.5 & 11.4 & 10.1 & 29.8 & 42.2 & 22.4 & 13.4 & 13.0 & 12.9 & 9.8 & 9.6 & 21.4 \\
 & \betae & 39.0 & {10.9} & {10.0} & 28.8 & {42.5} & {22.4} & {12.6} & {12.4} & 11.1 & {9.7} & \textbf{9.9} & {20.9} \\
 & \qb & {40.6} & 9.4 & 6.8 & {29.5} & 42.3 & 21.2 & {12.6} & 11.3 & - & 7.6 & - & 20.1 \\
 & \gqe & 35.0 & 7.2 & 5.3 & 23.3 & 34.6 & 16.5 & 10.7 & 8.2 & - & 5.7 & - & 16.3 \\ \midrule
\multirow{1}{*}{NELL995}
 & \logicept &  \textbf{58.3} & \textbf{17.7} & \textbf{15.4} & 40.5 & \textbf{50.4} & \textbf{27.3} & \textbf{19.2} & \textbf{15.9} & \textbf{14.4} & \textbf{12.7} & \textbf{11.8} & \textbf{28.6} \\
 & \logiceb & 57.6 & 16.9 & 14.6 & \textbf{40.8} & 50.2 & 26.5 & 18.0 & 15.3 & 13.5 & 12.0 & 11.3 & 28.0 \\
 & \betae & {53.0} & 13.0 & {11.4} & {37.6} & {47.5} & {24.1} & 14.3 & {12.2} & 11.0 & 8.5 & 8.6 & {24.6} \\
 & \qb & 42.2 & {14.0} & 11.2 & 33.3 & 44.5 & 22.4 & {16.8} & 11.3 & - & {10.3} & - & 22.9 \\
 & \gqe & 32.8 & 11.9 & 9.6 & 27.5 & 35.2 & 18.4 & 14.4 & 8.5 & - & 8.8 & - & 18.6 \\ \bottomrule
\end{tabular}
\end{small}
\end{center}
\vskip -0.1in
\end{table*}
\begin{table*}[!htpb]
\caption{Spearman's rank correlation (higher better) between learned embedding (diff. entropy and truth intervals for \logice, diff. entropy for \betae, box size for \qb) and the number of answers of queries.}
\label{tab:spearman}
\vskip 0.15in
\begin{center}
\begin{small}
\begin{tabular}{l|l|ccc|cc|cc|ccccc|c}
\toprule
\textbf{Dataset} & \textbf{Model} & \textbf{1p} & \textbf{2p} & \textbf{3p} & \textbf{2i} & \textbf{3i} & \textbf{pi} & \textbf{ip} & \textbf{2in} & \textbf{3in} & \textbf{inp} & \textbf{pin} & \textbf{pni} & \textbf{avg} \\
\midrule
\multirow{1}{*}{FB15k}
 & \logice entropy & \textbf{0.50} & \textbf{0.65} & \textbf{0.70} & 0.52 & {0.33} & \textbf{0.56} & \textbf{0.58} & \textbf{0.71} & \textbf{0.55} & \textbf{0.58} & \textbf{0.62} & \textbf{0.70} & \textbf{0.58} \\
 & \logice bounds & 0.44 & 0.51 & 0.50 & \textbf{0.60} & \textbf{0.52} & {0.58} & {0.48} & 0.56 & 0.53 & 0.33 & 0.43 & 0.59 & 0.51 \\
 & \betae & 0.37 & 0.48 & 0.47 & 0.57 & 0.40 & 0.52 & 0.42 & 0.62 & 0.55 & 0.46 & 0.47 & 0.61 & 0.49 \\
 & \qb & 0.30 & 0.22 & 0.26 & 0.33 & 0.27 & 0.30 & 0.14 & - & - & - & - & - & - \\
\midrule
\multirow{1}{*}{FB15k-237}
 & \logice entropy & \textbf{0.65} & \textbf{0.67} & \textbf{0.72} & 0.61 & 0.51 & \textbf{0.57} & \textbf{0.60} & \textbf{0.69} & 0.54 & \textbf{0.62} & \textbf{0.61} & \textbf{0.67} & \textbf{0.62} \\
 & \logice bounds & 0.61 & 0.58 & 0.58 & \textbf{0.64} & \textbf{0.64} & 0.54 & 0.49 & 0.58 & 0.50 & 0.41 & 0.49 & 0.60 & 0.56 \\
 & \betae & 0.40 & 0.50 & 0.57 & 0.60 & 0.52 & 0.54 & 0.44 & 0.69 & \textbf{0.58} & 0.51 & 0.47 & 0.67 & 0.54 \\
 & \qb & 0.18 & 0.23 & 0.27 & 0.35 & 0.44 & 0.36 & 0.20 & - & - & - & - & - & - \\
\midrule
\multirow{1}{*}{ NELL995}
 & \logice entropy & \textbf{0.67} & \textbf{0.64} & \textbf{0.60} & 0.63 & 0.64 & 0.55 & \textbf{0.58} & 0.70 & 0.59 & \textbf{0.49} & \textbf{0.55} & \textbf{0.70} & \textbf{0.61} \\
 & \logice bounds & 0.51 & 0.56 & 0.48 & \textbf{0.68} & \textbf{0.67} & \textbf{0.62} & 0.40 & 0.56 & \textbf{0.61} & 0.30 & 0.34 & 0.58 & 0.53 \\
 & \betae & 0.42 & 0.55 & 0.56 & 0.59 & 0.61 & 0.60 & 0.54 & \textbf{0.71} & 0.60 & 0.35 & 0.45 & 0.64 & 0.55 \\
 & \qb & 0.15 & 0.29 & 0.31 & 0.38 & 0.41 & 0.36 & 0.35 & - & - & - & - & - & - \\
\bottomrule
\end{tabular}
\end{small}
\end{center}
\vskip -0.1in
\end{table*}
\begin{table*}[!htpb]
\caption{Pearson correlation coefficient (higher better) between learned embedding (diff. entropy for \logice and \betae, box size for \qb) and the number of answers of queries.}
\label{tab:pearson}
\vskip 0.15in
\begin{center}
\begin{small}
\begin{tabular}{l|l|ccc|cc|cc|ccccc|c}
\toprule
\textbf{Dataset} & \textbf{Model} & \textbf{1p} & \textbf{2p} & \textbf{3p} & \textbf{2i} & \textbf{3i} & \textbf{pi} & \textbf{ip} & \textbf{2in} & \textbf{3in} & \textbf{inp} & \textbf{pin} & \textbf{pni} & \textbf{avg} \\
\midrule
\multirow{1}{*}{FB15k}
 & \logice & \textbf{0.28} & \textbf{0.50} & \textbf{0.56} & \textbf{0.47} & \textbf{0.34} & \textbf{0.38} & \textbf{0.43} & \textbf{0.56} & \textbf{0.46} & \textbf{0.45} & \textbf{0.48} & \textbf{0.56} & \textbf{0.46} \\
%  & \betae & 0.180 & 0.353 & 0.379 & 0.370 & 0.296 & 0.315 & 0.316 & 0.413 & 0.386 & 0.337 & 0.335 & 0.400 & 0.340 \\ % own results
 & \betae & 0.22 & 0.36 & 0.38 & 0.39 & 0.30 & 0.31 & 0.31 & 0.44 & 0.41 & 0.34 & 0.36 & 0.44 & 0.36 \\
 & \qb & 0.08 & 0.22 & 0.26 & 0.29 & 0.23 & 0.25 & 0.13 & - & - & - & - & - & - \\
\midrule
\multirow{1}{*}{FB15k-237}
 & \logice & \textbf{0.33} & \textbf{0.53} & \textbf{0.61} & \textbf{0.45} & 0.37 & \textbf{0.37} & \textbf{0.47} & \textbf{0.58} & \textbf{0.44} & \textbf{0.52} & \textbf{0.49} & \textbf{0.57} & \textbf{0.48} \\
%  & \betae & 0.196 & 0.358 & 0.417 & 0.357 & 0.293 & 0.325 & 0.337 & 0.431 & 0.380 & 0.366 & 0.338 & 0.428 & 0.352 \\ % own results
 & \betae & 0.23 & 0.37 & 0.45 & 0.36 & 0.31 & 0.32 & 0.33 & 0.46 & 0.41 & 0.39 & 0.36 & 0.48 & 0.37 \\
 & \qb & 0.02 & 0.19 & 0.26 & 0.37 & \textbf{0.49} & 0.34 & 0.20 & - & - & - & - & - & - \\
\midrule
\multirow{1}{*}{ NELL995}
 & \logice & \textbf{0.43} & \textbf{0.53} & \textbf{0.53} & \textbf{0.53} & \textbf{0.49} & \textbf{0.46} & \textbf{0.45} & \textbf{0.66} & \textbf{0.54} & \textbf{0.46} & \textbf{0.55} & \textbf{0.63} & \textbf{0.52} \\
%  & \betae & 0.233 & 0.421 & 0.394 & 0.378 & 0.345 & 0.387 & 0.287 & 0.475 & 0.474 & 0.261 & 0.321 & 0.447 & 0.369 \\ % own results
 & \betae & 0.24 & 0.40 & 0.43 & 0.40 & 0.39 & 0.40 & 0.40 & 0.52 & 0.51 & 0.26 & 0.35 & 0.46 & 0.40 \\
 & \qb & 0.07 & 0.21 & 0.31 & 0.36 & 0.29 & 0.24 & 0.34 & - & - & - & - & - & - \\
\bottomrule
\end{tabular}
\end{small}
\end{center}
\vskip -0.1in
\end{table*}
\begin{table*}[!htpb]
\caption{Answer size prediction mean absolute error (\%, lower better) with embedding entropy components for \logice and \betae, and box size components for \qb.}
\label{tab:sizepredict0}
\vskip 0.15in
\begin{center}
\begin{small}
\begin{tabular}{l|l|ccc|cc|cc|ccccc|c}
\toprule
\textbf{Dataset} & \textbf{Model} & \textbf{1p} & \textbf{2p} & \textbf{3p} & \textbf{2i} & \textbf{3i} & \textbf{pi} & \textbf{ip} & \textbf{2in} & \textbf{3in} & \textbf{inp} & \textbf{pin} & \textbf{pni} & \textbf{avg} \\
\midrule
\multirow{1}{*}{FB15k}
 & \logice & \textbf{78} & \textbf{86} & \textbf{91} & \textbf{88} & \textbf{93} & 96 & 95 & \textbf{85} & \textbf{83} & \textbf{82} & \textbf{86} & \textbf{84} & \textbf{87} \\
 & \betae & 105 & 95 & 96 & 92 & 94 & \textbf{92} & \textbf{94} & 95 & 92 & 95 & 95 & 96 & 95 \\
 & \qb & 264 & 113 & 100 & 444 & 730 & 386 & 125 & - & - & - & - & - & - \\
\midrule
\multirow{1}{*}{FB15k-237}
 & \logice & \textbf{78} & \textbf{83} & \textbf{86} & \textbf{82} & \textbf{94} & \textbf{89} & \textbf{86} & \textbf{81} & \textbf{79} & \textbf{81} & \textbf{81} & \textbf{81} & \textbf{83} \\
 & \betae & 111 & 96 & 97 & 97 & 97 & 95 & 97 & 97 & 95 & 97 & 97 & 98 & 98 \\
 & \qb & 191 & 101 & 100 & 310 & 780 & 263 & 103 & - & - & - & - & - & - \\
\midrule
\multirow{1}{*}{NELL995}
 & \logice & \textbf{59} & \textbf{82} & \textbf{85} & \textbf{83} & \textbf{83} & \textbf{90} & \textbf{83} & \textbf{76} & 92 & \textbf{77} & \textbf{78} & \textbf{77} & \textbf{80} \\
 & \betae & 97 & 96 & 95 & 94 & 100 & 97 & 95 & 95 & \textbf{90} & 96 & 94 & 96 & 95 \\
 & \qb & 254 & 100 & 100 & 1 371 & 2 638 & 539 & 109 & - & - & - & - & - & - \\
\bottomrule
\end{tabular}
\end{small}
\end{center}
\vskip -0.1in
\end{table*}
\begin{table*}[!htpb]
\caption{Detailed Hits@3 results for all the Query2Box datasets.}
\label{tab:emql_full}
\vskip 0.15in
\begin{center}
\begin{small}
\begin{tabular}{cl|ccccc|cccc|c}
\toprule
 \multicolumn{2}{c|}{{\textbf{Generalization}}}   & \textbf{1p}   & \textbf{2p}   & \textbf{3p}   & \textbf{2i}   & \textbf{3i}   & \textbf{ip}   & \textbf{pi}   & \textbf{2u}   & \textbf{up}  & \textbf{avg}  \\ \midrule
\multicolumn{1}{c|}{FB15k} & \logicept & \textbf{81.0} & \textbf{51.9} & \textbf{46.3} & 62.5 & {73.2} & 28.4 & 47.8 & \textbf{65.3} & 37.6 & \textbf{54.9} \\
\multicolumn{1}{c|}{} & \logiceb & 
76.4 & 47.9 & 43.3 & 56.6 & 67.1 & 25.0 & 42.6 & 57.9 & 35.9 & 50.3 \\
\multicolumn{1}{c|}{} & EmQL & 42.4 & {50.2} & {45.9}& \textbf{63.7} & {70.0} & \textbf{60.7} & \textbf{61.4} & 9.0  & \textbf{42.6} & {49.5} \\ 
% \multicolumn{1}{c|}{}          & ~~- sketch      & 50.6  & 46.7  & 41.6  & 61.8  & 67.3  & 54.2  & 53.5  & 21.6  & 40.0 & 48.6 \\
\multicolumn{1}{c|}{} & \betae & 75.8 & 46.0 & 41.8 & 62.5 & \textbf{74.3} & 24.3 & 48.0 & 60.2 & 29.2 & 51.4 \\
\multicolumn{1}{c|}{} & \qb & {78.6} & 41.3 & 30.3 & 59.3 & 71.2 & 21.1 & 39.7 & {60.8} & 33.0 & 48.4 \\
% \multicolumn{1}{c|}{} & ~~$+d$=2000      & 54.3 & 32.0 & 27.0 & 35.5 & 50.7 & 13.7 & 27.0 & 44.1 & 26.3 & 34.5 \\
\multicolumn{1}{c|}{} & \gqe & 63.6 & 34.6 & 25.0 & 51.5 & 62.4 & 15.1 & 31.0 & 37.6 & 27.3 & 38.7 \\
\midrule

\multicolumn{1}{c|}{FB15k-237} & \logicept & 46.1 & 28.6 & 24.8 & 34.8 & 46.5 & 12.0 & 23.7 & \textbf{27.7} & 21.1 & 29.5 \\
\multicolumn{1}{c|}{} & \logiceb & 45.0 & 26.6 & 23.0 & 32.0 & 44.1 & 11.1 & 22.1 & 25.5 & 20.4 & 27.7 \\
\multicolumn{1}{c|}{} & EmQL & 37.7 & \textbf{34.9} & \textbf{34.3} & \textbf{44.3} & \textbf{49.4} & \textbf{40.8} & \textbf{42.3} & 8.7  & \textbf{28.2} & \textbf{35.8}\\
%  ~~$-$ sketch & 43.1 & 34.6 & 33.7 & 41.0 & 45.5 & 36.7 & 37.2 & 15.3 & \textbf{32.5} & 35.5 & 48.6 & \textbf{46.8}\\
\multicolumn{1}{c|}{} & \betae & 43.1 & 25.3 & 22.3 & 31.3 & 44.6 & 10.2 & 22.3 & 26.6 & 18.0 & 27.1 \\
\multicolumn{1}{c|}{} & \qb & \textbf{46.7} & 24.0   & 18.6 & 32.4 & 45.3 & 10.8 & 20.5 & {23.9} & 19.3 & 26.8 \\
 %  ~~$+d$=2000            & 37.2  & 20.7  & 19.4  & 22.6  & 37.1  & 9.7  & 16.8  & 20.0  & 17.8 & 22.4 & 34.5 & 23.4 \\
\multicolumn{1}{c|}{} & \gqe & 40.5 & 21.3 & 15.5 & 29.8 & 41.1 & 8.5  & 18.2 & 16.9 & 16.3 & 23.1 \\
\midrule

\multicolumn{1}{c|}{NELL995} & \logicept & \textbf{64.5} & 36.4 & 36.6 & 41.4 & 54.6 & 14.9 & 26.0 & \textbf{51.4} & 27.9 & 39.3 \\
\multicolumn{1}{c|}{} & \logiceb & 63.9 & 35.1 & 35.5 & 40.7 & 54.4 & 14.2 & 25.2 & 50.3 & 27.6 & 38.6 \\
\multicolumn{1}{c|}{} & EmQL & 41.5 & \textbf{40.4} & \textbf{38.6} & \textbf{62.9} & \textbf{74.5} & \textbf{49.8} & \textbf{64.8} & 12.6 & \textbf{35.8} & \textbf{46.8}\\
% \multicolumn{1}{c|}{} & ~~- sketch      & 48.3 & 39.5 & 35.2 & 57.2 & 69.0 & 48.0 & 59.9 & 25.9 & \textbf{38.2} & \textbf{46.8} \\
\multicolumn{1}{c|}{} & \betae & 58.7 & 29.6 & 30.7 & 36.1 & 50.0 & 11.0 & 22.9 & 44.1 & 21.1 & 33.8 \\
\multicolumn{1}{c|}{} & \qb & {55.5} & 26.6 & 23.3 & 34.3 & 48.0 & 13.2 & 21.2 & {36.9} & 16.3 & 30.6 \\
% \multicolumn{1}{c|}{}          & ~~$+d$=2000         & 49.1 & 22.1 & 17.5 & 21.4 & 39.9 & 8.9 & 17.2 & 26.4 & 8.1 & 23.4\\
\multicolumn{1}{c|}{} & \gqe & 41.8 & 23.1 & 20.5 & 31.8 & 45.4 & 8.1  & 18.8 & 20.0 & 13.9 & 24.8 \\

\toprule
 \multicolumn{2}{c|}{{\textbf{Entailment}}}   & \textbf{1p}   & \textbf{2p}   & \textbf{3p}   & \textbf{2i}   & \textbf{3i}   & \textbf{ip}   & \textbf{pi}   & \textbf{2u}   & \textbf{up}  & \textbf{avg}  \\ \midrule
\multicolumn{1}{c|}{FB15k} & \logicept & 88.4 & 64.0 & 57.9 & 70.8 & 80.6 & 41.0 & 59.0 & 76.6 & 51.0 & 65.5 \\
\multicolumn{1}{c|}{} & \logiceb & 82.9 & 57.3 & 51.9 & 62.5 & 73.0 & 34.1 & 51.5 & 67.2 & 45.5 & 58.4 \\
\multicolumn{1}{c|}{} & EmQL & \textbf{98.5} & \textbf{96.3} & \textbf{91.1} & \textbf{91.4} & \textbf{88.1} & \textbf{87.8} & \textbf{89.2} & \textbf{88.7} & \textbf{91.3} &  \textbf{91.4}\\ 
\multicolumn{1}{c|}{}          & ~~$-$ sketch      & 85.1 & 50.8 & 42.4 & 64.4 & 66.1 & 50.4 & 53.8 & 43.2 & 42.7 & 55.5 \\
\multicolumn{1}{c|}{} & \betae & 83.2 & 57.3 & 51.0 & 71.1 & 81.4 & 32.7 & 56.9 & 70.4 & 41.0 & 60.6 \\
\multicolumn{1}{c|}{} & \qb & 68.0 & 39.4 & 32.7 & 48.5 & 65.3 & 16.2 & 32.9 & 61.4 & 28.9 & 43.7  \\
% \multicolumn{1}{c|}{}          & ~~$+d$=2000         & 59.0 & 36.8 & 30.2 & 40.4 & 57.1 & 14.8 & 28.9 & 49.2 & 28.7 & 38.3  \\
\multicolumn{1}{c|}{} & \gqe & 73.8 & 40.5 & 32.1 & 49.8 & 64.7 & 18.9 & 36.1 & 47.2 & 30.4 & 43.7 \\
\midrule

\multicolumn{1}{c|}{FB15k-237} & \logicept & 81.5 & 54.2 & 46.0 & 58.1 & 67.1 & 28.5 & 44.0 & 66.6 & 40.8 & 54.1 \\
\multicolumn{1}{c|}{} & \logiceb & 73.7 & 46.4 & 38.9 & 49.8 & 61.5 & 22.0 & 37.2 & 54.6 & 35.1 & 46.6 \\
\multicolumn{1}{c|}{} & EmQL & \textbf{100.0}  & \textbf{99.5} & \textbf{94.7} & \textbf{92.2} & \textbf{88.8} & \textbf{91.5} & \textbf{93.0} & \textbf{94.7} & \textbf{93.7} & \textbf{94.2} \\ 
\multicolumn{1}{c|}{} & ~~$-$ sketch & 89.3 & 55.7 & 39.9 & 62.9 & 63.9 & 51.9 & 54.7 & 53.8 & 44.7 & 57.4 \\
\multicolumn{1}{c|}{} & \betae & 77.9 & 52.6 & 44.5 & 59.0 & 67.8 & 23.5 & 42.2 & 63.7 & 35.1 & 51.8 \\
\multicolumn{1}{c|}{} & \qb & 58.5 & 34.3 & 28.1 & 44.7 & 62.1 & 11.7 & 23.9 & 40.5 & 22.0 & 36.2 \\
% ~~+$d$=2000            & 50.7 & 30.1 & 26.1 & 34.8 & 55.2 & 11.4 & 20.6 & 32.8 & 21.5 & 31.5 & 38.3 & 43.7\\
\multicolumn{1}{c|}{} & \gqe & 56.4 & 30.1 & 24.5 & 35.9 & 51.2 & 13.0 & 25.1 & 25.8 & 22.0 & 31.6 \\
\midrule

\multicolumn{1}{c|}{NELL995} & \logicept & 96.2 & 90.7 & 84.1 & 84.1 & 89.5 & 65.2 & 76.0 & 94.7 & 87.1 & 85.3 \\
\multicolumn{1}{c|}{} & \logiceb & 94.1 & 86.0 & 78.7 & 80.4 & 87.1 & 53.6 & 68.5 & 90.9 & 81.2 & 80.1 \\
\multicolumn{1}{c|}{} & EmQL & \textbf{99.0} & \textbf{99.0} & \textbf{97.1} & \textbf{99.7} & \textbf{99.6} & 
\textbf{98.7} & \textbf{98.9} & \textbf{98.8} & \textbf{98.5} & \textbf{98.8}\\
\multicolumn{1}{c|}{} & ~~$-$ sketch      & 94.5 & 77.4 & 52.9 & 97.4 & 97.5 & 88.1 & 90.8 & 70.4 & 73.5 & 82.5 \\
\multicolumn{1}{c|}{} & \betae & 94.3 & 88.2 & 76.2 & 84.0 & 90.2 & 46.6 & 68.8 & 92.5 & 81.4 & 80.2 \\
\multicolumn{1}{c|}{} & \qb & 83.9 & 57.7 & 47.8 & 49.9 & 66.3 & 19.9 & 29.6 & 73.7 & 31.0 & 51.1 \\
% \multicolumn{1}{c|}{} & ~~$+d$=2000         & 75.7 & 49.9 & 36.9 & 40.5 & 60.1 & 17.1 & 25.6 & 63.5 & 24.4 & 43.7 \\
\multicolumn{1}{c|}{} & \gqe & 72.8 & 58.0 & 55.2 & 45.9 & 57.3 & 24.8 & 34.2 & 59.0 & 40.7 & 49.8 \\
\bottomrule
\end{tabular}
\end{small}
\end{center}
\vskip -0.1in
\end{table*}

\end{document}